\documentclass[acmsmall]{acmart}
\AtBeginDocument{%
  \providecommand\BibTeX{{%
    \normalfont B\kern-0.5em{\scshape i\kern-0.25em b}\kern-0.8em\TeX}}}

\setcopyright{acmcopyright}
\copyrightyear{2026}
\acmYear{2026}
\acmDOI{XXXXXXX.XXXXXXX}

\acmJournal{JACM}
\acmVolume{37}
\acmNumber{4}
\acmArticle{111}
\acmMonth{5}

\usepackage{natbib}
\usepackage[utf8]{inputenc} 
\usepackage[T1]{fontenc}
\usepackage{amsthm}
\usepackage{tikz}
\usepackage[edges]{forest}
\usepackage{hyperref}
\usepackage{graphicx}
\usepackage{wrapfig}
\usepackage{subfigure}
\usepackage{multirow}
\usepackage{enumitem}
\usepackage{xcolor}
\usepackage{makecell}

\usepackage{array}
\newcolumntype{P}[1]{>{\raggedright\arraybackslash}p{#1}}
\newcolumntype{Y}{>{\raggedright\arraybackslash}X}

\definecolor{hiddendraw}{RGB}{205, 44, 36}
\definecolor{hidden-blue}{RGB}{194,232,247}
\definecolor{hidden-orange}{RGB}{243,202,120}
\definecolor{hidden-yellow}{RGB}{242,244,193}
\theoremstyle{definition}

\usepackage{tabularx}
\usepackage{booktabs}
\usepackage{pifont}
\newcommand{\cmark}{\ding{51}}%
\newcommand{\xmark}{\ding{55}}%

\begin{document}



\title{A Survey on Foundation Models for Personalized Federated Intelligence}

\author{Yu Qiao}
\email{qiaoyu1002@gmail.com}
\orcid{0000-0003-4045-8473}
\affiliation{%
  \institution{School of Computing, Kyung Hee University}
  \city{Yongin-si 17104}
  \country{Republic of Korea}
}

\author{Huy Q. Le}
\email{quanghuy69@khu.ac.kr}
\orcid{0009-0007-8342-7614}
\affiliation{%
  \institution{School of Computing, Kyung Hee University}
  \city{Yongin-si 17104}
  \country{Republic of Korea}
}

\author{Avi Deb Raha}
\email{avi@khu.ac.kr}
\orcid{0000-0003-0240-1214}
\affiliation{%
  \institution{School of Computing, Kyung Hee University}
  \city{Yongin-si 17104}
  \country{Republic of Korea}
}

\author{Phuong-Nam Tran}
\email{tpnam0901@khu.ac.kr}
\orcid{0009-0009-6551-9106}
\affiliation{%
  \institution{School of Computing, Kyung Hee University}
  \city{Yongin-si 17104}
  \country{Republic of Korea}}

\author{Apurba Adhikary}
\email{apurba@nstu.edu.bd}
\orcid{0000-0003-3970-1878}
\affiliation{%
 \institution{Noakhali Science and Technology University}
 \city{Noakhali-3814}
 \country{Bangladesh}}

\author{Mengchun Zhang}
\email{mengchun0527@gmail.com}
\orcid{0009-0008-0651-1932}
\affiliation{%
 \institution{Korea Advanced Institute of Science and Technology}
 \city{Daejeon}
 \country{Republic of Korea}}

\author{Loc X. Nguyen}
\email{xuanloc088@khu.ac.kr}
\orcid{0000-0001-5911-5847}
\affiliation{%
 \institution{School of Computing, Kyung Hee University}
 \city{Yongin-si 17104}
 \country{Republic of Korea}}

\author{Eui-Nam Huh}
\email{johnhuh@khu.ac.kr}
\orcid{0000-0003-0184-6975}
\affiliation{%
\institution{School of Computing, Kyung Hee University}
\city{Yongin-si 17104}
\country{Republic of Korea}}

\author{Dusit Niyato}
\email{dniyato@ntu.edu.sg}
\affiliation{%
  \institution{College of Computing and Data Science, Nanyang Technological University}
  \country{Singapore}}

\author{Choong Seon Hong}
\email{cshong@khu.ac.kr}
\orcid{0000-0003-3484-7333}
\affiliation{%
  \institution{School of Computing, Kyung Hee University}
  \city{Yongin-si 17104}
  \country{Republic of Korea}}

\renewcommand{\shortauthors}{Y. Qiao et al.}

\begin{abstract}
The rise of large language models (LLMs), such as ChatGPT, Gemini, and Grok, has reshaped the AI landscape. As prominent instances of foundational models (FMs), they exhibit remarkable capabilities in generating human-like content, pushing the boundaries towards artificial general intelligence (AGI). However, their large-scale nature, privacy sensitivity, and substantial computational demands pose significant challenges for personalized customization for end users. To bridge this gap, we present the vision of artificial personalized intelligence (API), which focuses on adapting FMs to individual users while ensuring privacy. As a central enabler of API, we propose personalized federated intelligence (PFI), a new paradigm that not only integrates the privacy benefits of federated learning (FL) with the generalization capabilities of FMs but also places personalization at its core. To this end, we first survey recent advances in FL and FMs that lay the foundation for PFI. We then explore core stages of the PFI pipeline: efficient personalization at the edge, trustworthy adaptation, and adaptive refinement via retrieval-augmented generation. Finally, we highlight future directions for enabling PFI. Overall, this survey aims to lay a foundation for the development of API as a complementary direction to AGI, with PFI as a key enabling paradigm.

\end{abstract}





\maketitle

\section{Introduction}
The landscape of AI has been profoundly transformed by the rise of foundation models (FMs), particularly large language models (LLMs) in natural language processing, such as ChatGPT~\cite{achiam2023gpt}, Gemini~\cite{team2023gemini}, and DeepSeek~\cite{qiao2025deepseek}. The advent of these LLMs has revolutionized tasks~\cite{gao2024examining} such as content generation, text summarization, and information retrieval~, showcasing their remarkable potential in understanding and generating human-like conversational responses. Building on this momentum, Meta AI released the segment anything model (SAM)~\cite{kirillov2023segment}, which claims to accurately segment any object in any image with remarkable zero-shot generalization for downstream tasks. This breakthrough, regarded as the "ChatGPT moment" for computer vision, has inspired numerous follow-up works aimed at further enhancing SAM’s capabilities, including lightweight model designs~\cite{zhang2024efficientvit} and fine-tuning for medical image analysis~\cite{ma2024segment}. Beyond the image domain, Meta AI recently took a significant step forward by introducing SAM2~\cite{ravi2024sam}, which extends the promptable capabilities of SAM to the video domain, enabling fast and precise object selection in any video or image. Subsequent studies have further adapted SAM2 for various downstream tasks, such as 3D medical image segmentation~\cite{he2025few}. The success of these FMs has bolstered the anticipation that the era of strong AI is approaching, bringing us closer to the realization of artificial general intelligence. \textcolor{black}{List of abbreviations and definitions is provided to facilitate the smooth flow of the paper in Table~\ref{Notation}.}

\begin{table*}[t]
\centering
\color{black}
\caption{\textcolor{black}{List of Abbreviations and Their Definitions.}}
\label{Notation}
\vspace{-10px}
\resizebox{0.95\textwidth}{!}{
\begin{tabular}{p{1.4cm}|p{3.4cm}|p{1.4cm}|p{3.4cm}|p{1.4cm}|p{3.4cm}}
\toprule
\textbf{Abbr.} & \textbf{Definitions} & \textbf{Abbr.} & \textbf{Definitions} & \textbf{Abbr.} & \textbf{Definitions}\\ \hline

AGI & Artificial General Intelligence & API & Artificial Personalized Intelligence &  FMs & Foundation Models  \\ \hline
PFI & Personalized Federated Intelligence &FL &Federated Learning & LLMs & Large Language Models\\ \hline
SAM & Segment Anything Model &FFMs &Federated Foundation Models & RL  & Reinforcement Learning \\ \hline
RLHF &  Reinforcement Learning from Human Feedback & RLAIF &  Reinforcement Learning from AI Feedback & RAG  & Retrieval-Augmented Generation \\ \hline
LoRA &  Low-rank Adaptation & PEFT & Parameter-efficient fine-tuning  & XAI  & Explainable AI\\ 

\bottomrule
\end{tabular}}
\vspace{-13px}
\end{table*}
\color{black}

However, FMs are widely known for their massive model sizes and reliance on extensive and diverse datasets sourced from books, web pages, and other domains for training~\cite{zhou2024comprehensive}. As a result, they inherently face limitations, including challenges related to data privacy, computational resource constraints, and limited adaptability to domain-specific tasks without further fine-tuning~\cite{lin2023pushing}. For instance, training a GPT-3 model with 175 billion parameters requires 1,024 Nvidia V100 GPUs running for 34 days at an estimated cost of \$4.6 million US dollars~\cite{zhao2024optimizing}. Such high barriers make it infeasible for smaller organizations or emerging companies to develop or even fine-tune such large-scale models, particularly under limited computational and financial constraints. In addition, end users are often reluctant to share private data for training and are concerned about its potential misuse by large models like GPT~\cite{wu2024unveiling}. The training datasets themselves may also lack global representativeness, leading to embedded biases. For instance, associating men with engineering roles and women with caregiving tasks, or favoring majority racial groups over minorities in job recommendations~\cite{lyu2023pathway}. Moreover, data in the real world is continuously evolving~\cite{aggarwal2023chameleon}, making it challenging to keep large models up-to-date. Retraining or even fine-tuning these models with incremental data is highly costly and poses significant difficulties for end users who lack the resources or expertise to modify them. Finally, personalized applications require models that can adapt to individual users' preferences and unique data patterns. For instance, a healthcare platform may need to offer tailored diagnoses based on a user's medical history, while an e-commerce recommendation system should adjust to personalized shopping behaviors. Therefore, traditional centralized FM training struggles to meet these needs, as it needs to balance AGI ambitions with the requirement for privacy-preserving personalized intelligence.

These limitations highlight the urgent need for a new paradigm that enables FMs to be both powerful and personally adaptive, while respecting users’ privacy and resource constraints. To address this, we propose a new vision, \textbf{artificial personalized intelligence (API)}, which complements and extends the pursuit of AGI: whereas AGI targets universal capabilities across tasks and users, API focuses on intelligence tailored to individual needs. To realize this vision, we introduce \textbf{personalized federated intelligence (PFI)}, a key enabling paradigm grounded in federated foundation models (FFMs), which are FMs collaboratively trained via FL across decentralized data silos. At its core, PFI emphasizes personalization, aiming to deliver user-adaptive intelligence while preserving privacy. Note that FL is a decentralized training framework that allows models to be collaboratively trained across distributed entities without sharing raw data, thereby supporting privacy protection, data sovereignty, and regulatory compliance. This makes it particularly suitable for privacy-sensitive domains such as healthcare, finance, and personalized services.

\begin{table}[t]
\centering
\caption{\textcolor{black}{Comparison of Key Concepts: FL, FM, FFM, AGI, API, and PFI.}}
\label{tab:concepts}
\vspace{-10px}
\scalebox{0.80}{%
\begin{tabular}{p{0.8cm}|p{14cm}}
\toprule
\textbf{Abbr.}  & \textbf{Relation to \textcolor{green!60!black}{PFI}} \\ 
\midrule
\textcolor{black}{\textbf{FL}} & \textcolor{black}{
Serves as a foundational mechanism for enabling privacy-preserving training in \textcolor{green!60!black}{PFI}.} \\
\midrule
\textcolor{black}{\textbf{FM}}  & \textcolor{black}{Provides the generalization capability that \textcolor{green!60!black}{PFI} aims to personalize.} \\
\midrule
\textcolor{black}{\textbf{FFM}} & \textcolor{black}{Focuses on distributed training and adaptation of FMs; \textcolor{green!60!black}{PFI} builds on FFM principles but further highlights user-level personalization.} \\
\midrule
\textcolor{orange!80!black}{\textbf{AGI}} & \textcolor{orange!80!black}{A complementary vision to \textcolor{orange!80!black}{API}; aims for general intelligence, while API focuses on personalized intelligence.} \\
\midrule
\textcolor{orange!80!black}{\textbf{API}}  & \textcolor{orange!80!black}{The overarching goal of \textcolor{green!60!black}{PFI}, emphasizing user-specific intelligence rather than general-purpose intelligence.} \\
\midrule
\textcolor{green!60!black}{\textbf{PFI}}  & \textcolor{green!60!black}{The central paradigm proposed in this work, integrating \textcolor{black}{FL} and \textcolor{black}{FM} to realize privacy-preserving personalized intelligence.} \\
\bottomrule
\end{tabular}}
\vspace{-17px}
\end{table}

In the PFI framework, the global FM can either be a well-trained off-the-shelf model or be collaboratively trained from scratch across multiple decentralized data silos equipped with powerful computing resources (e.g., hospitals or enterprises), thereby ensuring privacy protection from the outset. This shared model is then efficiently adapted at the edge, enabling lightweight personalization aligned with local data distributions or user-specific preferences. Further enhancements can be introduced through trustworthy personalization (e.g., fairness-aware or robust optimization) and retrieval-augmented personalization (e.g., dynamic integration of external, user-relevant knowledge), allowing the model to continuously evolve with individual user contexts. Overall, this paper aims to explore the motivations, emerging solutions, and future directions of PFI, offering a comprehensive overview of the field and envisioning the path towards the API era.

\textcolor{black}{As the above discussion involves several interrelated but potentially complex concepts, Table~\ref{tab:concepts} provides a concise comparison of FL, FM, FFM, AGI, API, and PFI to help readers clearly distinguish their goals and the relation to our key proposal, PFI.} In summary, while we propose the conceptual framework of PFI, this work is structured as a survey that systematically reviews and categorizes existing literature under this emerging paradigm.

\begin{table*}[t]
    \centering
    \caption{\textcolor{black}{Comparison of Existing Survey and Position Papers with This Survey Paper.}}
    \label{tab_tease_example}
    \vspace{-10px}
    \renewcommand{\arraystretch}{1.1}
    \resizebox{0.95\linewidth}{!}{%
    \begin{tabular}{c|c|c c c c c c}
        \toprule
        Year & Ref. & FL Taxonomy & FM Taxonomy & Efficient FFM & Trustworthy FFM & Adaptive FFM & Personalization \\
        \midrule
        \multirow{3}{*}{2023}
        & ~\cite{zhuang2023foundation} & \xmark & \xmark & \cmark & \cmark & \xmark & \xmark \\
        & ~\cite{yu2023federated} & \xmark & \cmark & \cmark & \xmark & \xmark & \cmark \\
        & ~\cite{wen2023survey} & \cmark & \xmark & \xmark & \xmark & \xmark & \cmark \\
        \midrule
        \multirow{4}{*}{2024}
        & ~\cite{woisetschlager2024survey} & \xmark & \cmark & \cmark & \xmark & \xmark & \xmark \\
        & ~\cite{ren2025advances} & \xmark & \cmark & \cmark & \cmark & \xmark & \xmark \\
        & ~\cite{li2024synergizing} & \xmark & \cmark & \cmark & \cmark & \cmark & \xmark \\
        & ~\cite{li2024position} & \xmark & \cmark & \xmark & \cmark & \xmark & \xmark \\
        \midrule
        \multirow{6}{*}{2025}
        & ~\cite{jiang2025comprehensive} & \xmark & \cmark & \cmark & \cmark & \xmark & \xmark \\
        & ~\cite{kang2025grounding} & \xmark & \cmark & \cmark & \cmark & \xmark & \xmark \\
        & ~\cite{bian2025survey} & \cmark & \cmark & \cmark & \xmark & \xmark & \xmark \\
        & ~\cite{ye2025vertical} & \cmark & \cmark & \cmark & \cmark & \xmark & \xmark \\
        & ~\cite{xu2025resource} & \xmark & \cmark & \cmark & \xmark & \xmark & \xmark \\
        & \textcolor{black}{~\cite{fan2025ten}} & \textcolor{black}{\xmark} & \textcolor{black}{\cmark} & \textcolor{black}{\cmark} & \textcolor{black}{\cmark} & \textcolor{black}{\xmark} & \textcolor{black}{\xmark} \\
        \midrule
        \multicolumn{2}{c|}{\textbf{Ours}} & \cmark & \cmark & \cmark & \cmark & \cmark & \cmark\\
        \bottomrule
    \end{tabular}}
    \vspace{-16px}
\end{table*}

\subsection{Major Contributions}
This survey represents the first comprehensive exploration envisioning an era where API complements the widely discussed AGI. In contrast to prior surveys that primarily focus on specific aspects of FL and FMs, our work introduces a new paradigm, PFI, which emphasizes user-specific personalization and provides broader coverage of the topic. Table~\ref{tab_tease_example} compares our survey with several existing studies. While prior studies have provided valuable insights into dimensions such as efficiency, trustworthiness, or FM taxonomy, most of them examine these aspects in isolation rather than within an integrated conceptual framework. \textcolor{black}{For example, the work in~\cite{fan2025ten} summarize the ten major challenges of federated foundation models and provide grounded analyses of their objectives and potential solutions; however, they remain centered on the FM-centric perspective. In contrast, our survey takes a PFI-centric view, treating personalization as the core objective and structuring the design space along the entire PFI pipeline: efficient edge-side personalization, trustworthy adaptation, and adaptive refinement, thereby enabling continuous alignment with dynamic user-centric contexts. A more detailed comparison with existing works is provided in Appendix A.} We hope this roadmap paves the way towards the realization of truly personalized and privacy-preserving intelligence. Overall, the key contributions of this survey are as follows:
\begin{itemize} 
\item[$\bullet$] With the goal of advancing an API era that complements the development of AGI, this survey aims to shift focus towards the personalization aspect by exploring PFI, which, in contrast to prior work that has mainly concentrated on the role of FMs in enhancing the generalization capabilities of FL models, emphasizes the importance of end-user personalization.
\item[$\bullet$] To provide a clear understanding of both FL and FMs, we revisit the key enabling technologies and the taxonomy of FL and FMs, highlighting their synergies and identifying key gaps that motivate the integration of personalization into FFMs.
\item[$\bullet$] To lay the foundation for practical PFI deployment, we highlight several key technical directions, including efficient adaptation strategies for edge environments; mechanisms to ensure trustworthiness under real-world constraints such as privacy, security, and resource limitations; and adaptive design leveraging retrieval-augmented generation techniques.
\item[$\bullet$] Looking ahead, we outline a forward-looking research agenda encompassing Meta-PFI, quantum-enabled learning, and sustainable intelligence, aiming to inspire future exploration towards next-generation personalized FMs that are user-centric and well-suited for decentralized, privacy-sensitive environments.
\end{itemize}

\subsection{Organization of This Paper}
The remainder of this survey is organized as follows. Section~\ref{sec_background} provides an overview of FL and FMs, highlighting the motivation for PFI, where FFMs form the core framework but require additional techniques for effective personalization. Building on this foundation, Section~\ref{sec_efficient_PFI} discusses efficient personalization at the edge. Section~\ref{sec_trustworthy_PFI} discusses trustworthy adaptation. Section~\ref{sec_adaptive_PFI} explores further adaptive refinement. Section~\ref{sec_future} outlines future directions in this emerging field. Finally, Section~\ref{sec:conclusion} concludes the survey.

\section{Background}
\label{sec_background}
\subsection{Federated Learning}
Federated learning (FL)~\cite{mcmahan2017communication}, initially proposed by Google to enhance data privacy in Android ecosystems, enables collaborative model training across distributed devices without sharing raw data. This privacy-preserving paradigm has shown strong potential across domains such as healthcare (e.g., Tencent Tianyan Lab and WeBank’s medical FL framework~\cite{ju2020privacy}), finance (e.g., WeBank’s FATE for credit risk assessment~\cite{liu2021fate}), and advertising (e.g., JD’s 9N-FL framework boosting multi-party revenue~\cite{liu2024vertical}). More recently, FL has been applied to LLM training, as demonstrated by Prime Intellect~\cite{jaghouar2024intellect} and Photon~\cite{sani2024photon}. In addition, FL supports edge computing scenarios, including smart city applications like traffic management. These advancements highlight FL's growing role in both academic research and real-world deployment in the era of FMs.

The success of these applications stems from the scalable and privacy-preserving nature of FL, which enables model training across heterogeneous environments without sharing raw data~\cite{mcmahan2017communication}. \textcolor{black}{In a typical FL setup, each entity (e.g., financial institution) receives a global model from a central server, performs local training on private data, and sends updated parameters back to the server. The server aggregates these updates and redistributes the refined model to the entities for the next global iteration. Formally, the goal of 
standard FL is to learn a \emph{single global model} $\omega$ by minimizing
\begin{equation}
F(\omega) = \sum_{k=1}^{K} p_k F_k(\omega),
\end{equation}
where $F_k(\omega)$ is the local loss of client $k$ evaluated on its dataset 
$\mathcal{D}_k$, and $p_k = \frac{n_k}{\sum_j n_j}$.} To provide a clear understanding and taxonomy of the FL paradigm, we categorize FL from different perspectives. First, based on the types of participating entities, FL can be classified into cross-silo FL and cross-device FL. Second, considering the patterns of underlying data distribution across entities, FL can be categorized into horizontal FL, vertical FL, and federated transfer learning~\cite{yang2019federated}.

\begin{figure*}[t]
\centering
\includegraphics[width=0.80\textwidth]{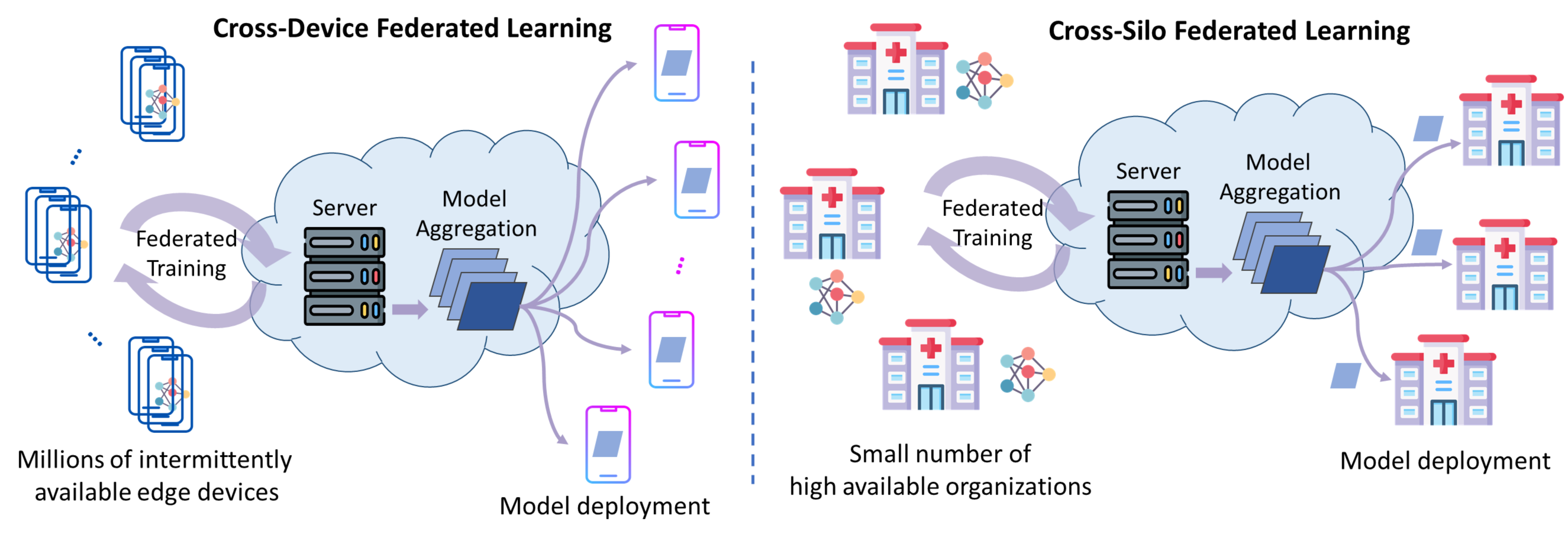}
\vspace{-10px}
\caption{Illustration of cross-device and cross-silo federated learning scenarios~\cite{kairouz2021advances}.}
\label{fig:cross_silo_device_fl}
\vspace{-15px}
\end{figure*}

\textbf{Participating Entities.} Based on the types of participating entities, the taxonomy of FL can be classified into two main categories: cross-silo FL and cross-device FL~\cite{kairouz2021advances}. Cross-silo FL involves a small number of organizational entities, such as banks, hospitals, or research institutes, where data remains local, and each entity typically has abundant computational resources. In contrast, cross-device FL involves numerous distributed personal devices, such as smartphones and IoT devices, which typically have limited computational resources. Figure~\ref{fig:cross_silo_device_fl} illustrates both scenarios. 

\begin{figure*}[t]
\centering
\subfigure[\textcolor{black}{Horizontal Federated Learning}]{
 \begin{minipage}[t]{0.30\linewidth}
 \centering
 \includegraphics[width=1\linewidth]{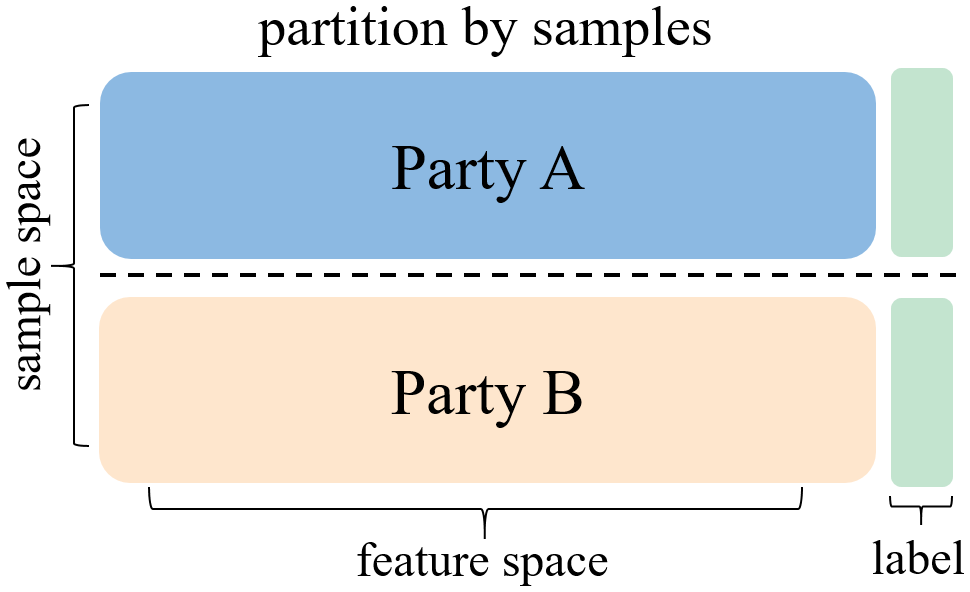}
 \end{minipage}
 }
 \subfigure[\textcolor{black}{Vertical Federated Learning}]{
 \begin{minipage}[t]{0.30\linewidth}
 \centering
 \includegraphics[width=1\linewidth]{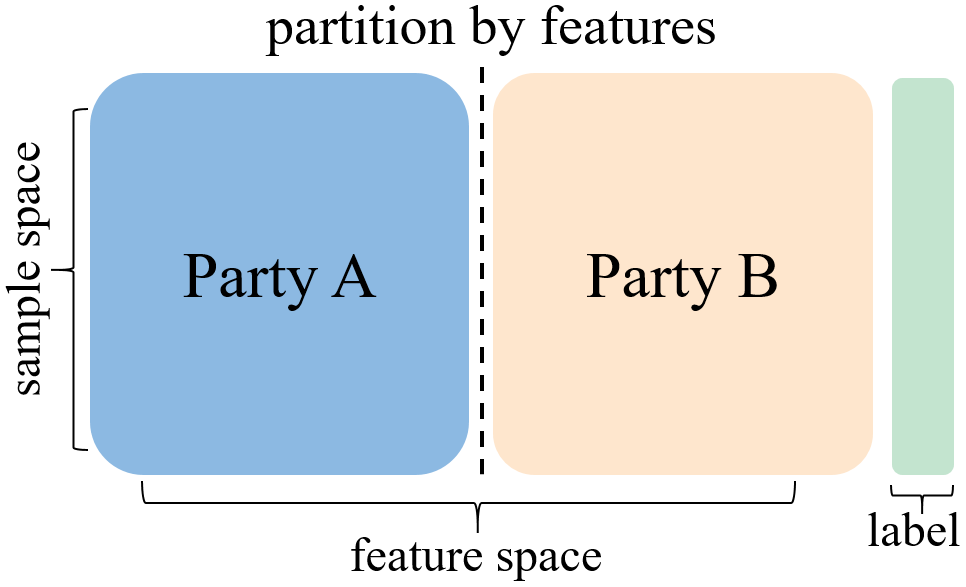}
 \end{minipage}
 }
 \subfigure[\textcolor{black}{Federated Transfer Learning}]{
 \begin{minipage}[t]{0.30\linewidth}
 \centering
 \includegraphics[width=1\linewidth]{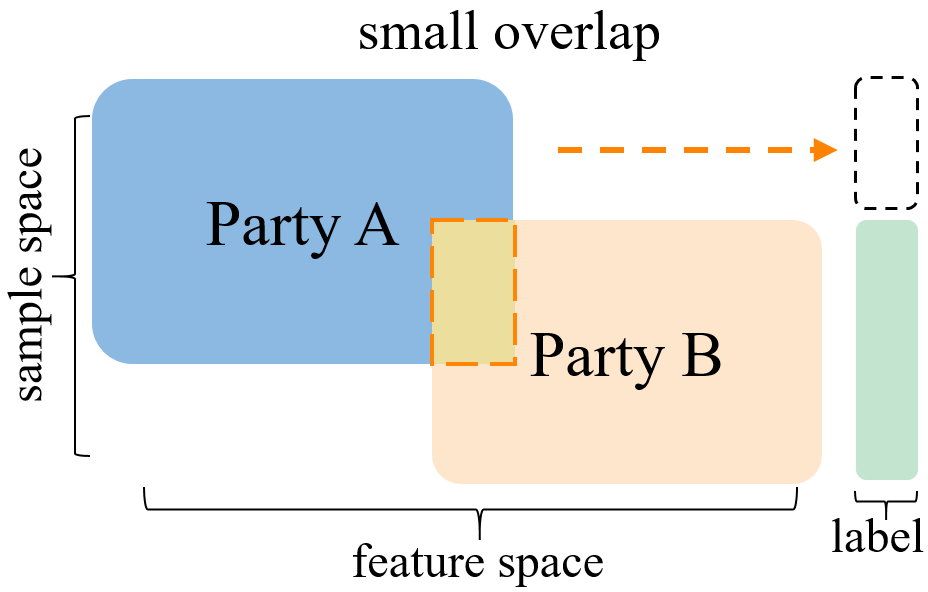}
 \end{minipage}
 }
 \vspace{-15px}
\caption{\textcolor{black}{Illustration of horizontal FL, vertical FL, and federated transfer learning~\cite{liu2024vertical}.}}    
\vspace{-20px}
 \label{fig:hfl_vfl_ftl}
\end{figure*}

\textbf{Data Patterns.} Based on the characteristics of the involved data distribution, the taxonomy of FL can be categorized into three main types: horizontal FL, vertical FL, and federated transfer learning. This taxonomy was also proposed in~\cite{liu2024vertical}. Here, we include it for a comprehensive overview and provide a more detailed explanation of the differences and applications of each type. Horizontal FL is most effective when participants share the same feature space but differ in their data samples, such as in collaborative model training across multiple hospitals. Vertical FL, on the other hand, involves entities with complementary features, such as the cooperation between banks and e-commerce platforms, where data about the same users is distributed across different feature spaces. For example, banks and e-commerce platforms can jointly train a model without sharing data, leveraging each other's insights: banks enhance credit risk assessments, while e-commerce platforms optimize marketing strategies and product recommendations. Finally, federated transfer learning is applied when participants have minimal overlap in both features and samples, such as in cross-domain tasks where data from different industries or regions are integrated to improve model performance. \textcolor{black}{Figure~\ref{fig:hfl_vfl_ftl} provides a visual representation of these types.}

\subsection{Overview of Foundation Models}

FMs~\cite{awais2025foundation}, also known as large X models, are deep learning models trained on large-scale, vast datasets to enable broad applicability across various domains~\cite{qiao2025deepseek}. Their development requires substantial computational resources and financial investment, but once trained, they can serve as a foundation for adaptation (e.g., fine-tuning) across a wide range of downstream tasks~\cite{bommasani2021opportunities}. Generative AI applications, such as LLMs, have emerged as key areas where FMs play a crucial role. The success of LLMs is largely attributed to the massive scaling of both data and model size, while foundational components like the Transformer~\cite{khan2022transformers} and RL~\cite{franccois2018introduction} have been established for many years. For instance, OpenAI's GPT series, such as GPT-3 with billions of parameters, has demonstrated impressive zero-shot learning abilities across tasks like question answering and commonsense reasoning~\cite{radford2018improving}. Similarly, xAI's Grok-3, with trillions of parameters, is regarded as one of the smartest AIs on Earth, showcasing state-of-the-art capabilities in solving a wide range of challenging problems, including mathematics, reasoning, and programming~\cite{jegham2025visual}. More recently, models such as DeepSeek-R1~\cite{guo2025deepseek} have achieved breakthroughs by significantly reducing training costs while maintaining performance comparable to OpenAI-o1-1217 on reasoning tasks. Notably, DeepSeek-R1 employs a pure RL-based training pipeline without any warm-up stage, marking a shift towards greater efficiency and accessibility in foundation model development. In contrast to the training objective in FL, which aims to minimize aggregated loss across multiple decentralized entities, FMs are typically trained in a centralized manner. The training objective of a foundation model can be expressed in a simplified form as:
\begin{equation}
\min_{\omega} \; \mathbb{E}_{x \sim \mathcal{D}} \left[ \mathcal{L}(F_\omega(x)) \right],
\vspace{-3px}
\end{equation}
where $\mathcal{L}$ denotes the selected loss function, and $F_\omega(x)$ represents the parameterized model to be optimized. Here, $x$ is the training samples drawn from a large-scale dataset $\mathcal{D}$. \textcolor{black}{In practice, however, FM training is not a single-stage process but rather a composition of multiple stages that operate sequentially and complementarily. Typically, a model is first pre-trained on massive unlabeled corpora using self-supervised objectives to capture general features and knowledge, providing a good initialization for downstream tasks. It is then fine-tuned on specific datasets to adapt to downstream tasks. Building on this, a further RL-based alignment stage is commonly applied, such as RL from human feedback (RLHF)~\cite{ouyang2022training}, RL from AI feedback (RLAIF)~\cite{bai2022constitutional}, and pure RL~\cite{guo2025deepseek}), to guide the model towards outputs that better align with human expectations and social norms. These stages form an integrated pipeline where pre-training establishes general capabilities, fine-tuning refines task adaptability, and RL aligns model behavior with human or AI preferences. }

To provide a systematic classification of the FM paradigm, we propose a taxonomy that organizes FMs across four key dimensions. First, based on input data modality, FMs can be classified into two types: unimodal models, which are trained on a single modality such as text (e.g., BERT~\cite{devlin2019bert}), vision (e.g., MAE~\cite{he2022masked}), or audio (e.g., Wav2Vec~\cite{schneider2019wav2vec}); and multimodal models, which are designed to process and integrate information from multiple modalities, such as image-text (e.g. CLIP~\cite{radford2021learning}) or video-language (e.g. VideoBERT~\cite{sun2019videobert}) pairs. Second, from a model architecture perspective, FMs can be categorized into three types: dense models, which activate all parameters during both training and inference (e.g., BERT)~\cite{riquelme2021scaling}; sparse models, such as mixture-of-experts (MoE), where only a subset of parameters is activated for each input to improve efficiency and scalability (e.g., GLaM~\cite{du2022glam}, DeepSeekMoE~\cite{dai2024deepseekmoe}); and retriever-augmented models, which leverage external memory or retrieval mechanisms to enhance factual accuracy and knowledge utilization (e.g., RETRO~\cite{borgeaud2022improving} and retrieval augmented generation (RAG)~\cite{lewis2020retrieval}). Third, in terms of training paradigms, FMs can be broadly categorized into two groups. The first category comprises self-supervised pretraining models, which learn general-purpose representations from large-scale unlabeled data by solving pretext tasks such as masked token prediction or contrastive learning (e.g., MAE~\cite{he2022masked} and BERT~\cite{devlin2019bert}). The second category encompasses RL-based training strategies, including RLHF~\cite{ouyang2022training}, RLAIF~\cite{bai2022constitutional}, and pure RL~\cite{guo2025deepseek} approaches. These methods aim to improve alignment with user intent and enhance the safety and robustness of models in real-world applications. For instance, InstructGPT is trained using RLHF, where human preference data serves as a reward signal to fine-tune the model’s behavior~\cite{ouyang2022training}. Claude leverages RLAIF, in which a preference model is used as the reward function without any human labels~\cite{bai2022constitutional}. DeepSeek-R1-Zero, on the other hand, is trained via large-scale pure RL without any supervised fine-tuning stages such as RLHF or RLAIF, serving as an early exploration into instruction-following driven purely by RL~\cite{guo2025deepseek}. Finally, regarding deployment and usage, FMs can be classified into two broad categories: cloud-centric models, which are typically deployed on centralized infrastructure and accessed via APIs; and on-device models, which are optimized for local inference, personalization, and privacy-preserving applications. For instance, GPT-4 and Claude are cloud-based models that offer powerful capabilities through API access, enabling large-scale applications across various domains. In contrast, models such as Meta’s on-device TinyLLaMA~\cite{zhang2024tinyllama} and Google’s Gemini Nano~\cite{team2023gemini} are specifically designed to run directly on edge devices, such as smartphones, offering reduced latency and improved data privacy. \textcolor{black}{This taxonomy provides a systematic framework for analyzing the capabilities, design trade-offs, and deployment scenarios of FMs, with a comparative overview presented in Appendix B.}

\subsection{Motivation for Personalized Federated Intelligence}

\textcolor{black}{Complementary to AGI, which aims to achieve general intelligence with universal capabilities across tasks and users, this survey envisions an era of API that emphasizes intelligence tailored to individual needs and contexts.  Realizing such personalization, however, requires access to diverse user data for model training, which poses significant privacy and governance challenges under traditional centralized paradigms. Centralized strategies that aggregate data in a single location risk violating privacy regulations and exacerbating issues related to data and computing resource monopolies. In this context, FL~\cite{mcmahan2017communication}, with its inherent privacy-preserving and decentralized nature, serves as a key enabler of large-scale collaboration across users and organizations while ensuring the protection of sensitive data. Moreover, based on its distributed learning paradigm, FL enables collaboration among multiple devices, thereby overcoming the limitations imposed by the scale and diversity of data available on any single device. Empirical evidence further supports the value of such collaboration. In particular, FATE-LLM\cite{fan2023fate} demonstrates that federated LoRA and federated P-Tuning-v2 generally outperform individual clients’ individual local fine-tuning across ROUGE and BLEU metrics. Notably, this improvement is achieved while communicating only a small fraction of parameters, 0.058\% for LoRA and 0.475\% for P-Tuning-v2, relative to full fine-tuning. Similarly, WorldLM~\cite{iacob2024worldwide} shows that personalized federated training, enabled by partial model localization and adaptive cross-federation knowledge sharing, achieves up to 1.91× performance improvements over standard FL, while closely approaching the performance of fully local personalized models under privacy-enhancing constraints. Nonetheless, FL faces fundamental challenges in adapting to the diverse data distributions and resource constraints across participants~\cite{le2024cross}. In real-world scenarios, participant data is often non-independent and identically distributed, as each user or device typically exhibits unique characteristics influenced by factors such as geographic location, individual behavior, and personal preferences~\cite{sabah2025fairdpfl}. For instance, typing behaviors on mobile keyboards differ across users, smartwatch health data varies with individual physiology and routines, and corporate logs capture unique operational patterns across organizations~\cite{sivakumar2024emg2qwerty}. Such heterogeneity poses a major challenge for traditional FL methods, which optimize a single global model that may fail to capture user diversity~\cite{li2020federated}. This limitation motivates the need for PFI, where models are collaboratively trained while being locally adapted to align with individual data distributions. Moreover, given the growing availability of lightweight FMs (e.g., OpenAI o1-mini~\cite{wang2024ontheplanning}, Lite-SAM~\cite{fu2024lite}, and DeepSeek-V2-Lite~\cite{liu2024deepseek}), PFI offers a practical pathway to efficiently fine-tune and adapt these models across diverse user environments with minimal communication and computational overhead.}

\textcolor{black}{\textbf{Personalized Federated Intelligence (PFI).} Unlike classical FL, which aims to learn a single global model, PFI seeks to learn a \emph{set of personalized models}, $\Omega=\{\omega_1,\dots,\omega_K\}$, tailored to the heterogeneous data distributions of $K$ participants. In general, each client model can be decomposed into two components, formulated as follows:
\begin{equation}
\vspace{-2px}
\omega_k = \Phi(\omega_g, \theta_k),
\end{equation}
where $\omega_g$ denotes the shared global parameters, $\theta_k$ denotes client-specific parameters, and $\Phi(\cdot)$ is the personalization mapping that integrates global and local components. }

\textcolor{black}{Under this characterization, the PFI objective can be formulated as follows:
\begin{equation}
\vspace{-2px}
    \min_{\omega_g,\,\Theta}
    \sum_{k=1}^{K} p_k\, F_k\!\left(\Phi(\omega_g, \theta_k)\right),
\end{equation}
where $\Theta=\{\theta_1,\dots,\theta_K\}$ and $F_k(\cdot)$ is the local loss on $\mathcal{D}_k$. This formulation provides a rigorous distinction from standard FL, which optimizes only $\omega_g$, by jointly learning shared and personalized components. }

\textcolor{black}{For completeness, the general objective can also be written directly over personalized models, as follows:
\begin{equation}
\vspace{-2px}
    \min_{\Omega}\; \sum_{k=1}^{K} p_k\, F_k(\omega_k),
\end{equation}
where each $\omega_k$ is generated through the personalization mechanism above.} \textcolor{black}{In practice, participants only fine-tune a subset of parameters, such as adapters or LoRA modules (introduced in the following sections), rather than the entire model, to reduce communication and computational costs, enabling PFI to achieve personalized performance at a substantially lower cost than full-model training, making it more practical and scalable.}

\section{Towards Efficient Personalized Federated Intelligence}
\label{sec_efficient_PFI}

\textcolor{black}{As the foundational stage of PFI, efficient personalization enables scalable deployment and lightweight adaptation of FMs for resource-constrained clients. Given the massive scale of FMs, directly deploying them is often impractical. Therefore, lightweight model designs are crucial for practical personalization. This section first establishes the efficiency-oriented foundation of PFI, which subsequently supports trustworthy deployment and continuous adaptive refinement in later stages.}

\subsection{Communication-and Computation-Efficient Strategies}

\begin{wrapfigure}{r}{0.45\textwidth} 
  \centering
  \includegraphics[width=0.4\textwidth]{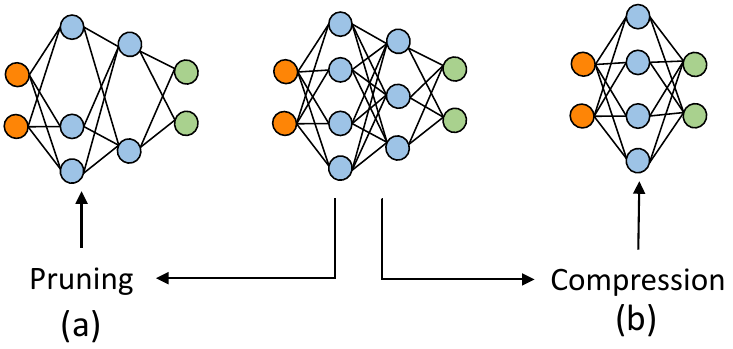}
  \vspace{-12px}
  \caption{\textcolor{black}{Illustration of model pruning and model compression. (a) Model pruning removes redundant components (e.g., neurons or layers) to reduce computation and communication. (b) Model compression encodes parameters for compact transmission and storage.}}
  \label{fig:pruning_compression}
    \vspace{-13px}
\end{wrapfigure}

Deploying FMs in FL environments introduces significant communication and computational bottlenecks due to the massive size of model parameters and limitations of edge devices~\cite{chen2024feddat}. To address these challenges, recent research has explored lightweight model adaptation techniques, such as model pruning and compression, to reduce both the volume of transmitted information and the computational overhead during training and inference.

\textbf{Model Pruning.} Among these, model pruning eliminates a model's redundant or less important components, such as neurons, attention heads, or even entire layers, to create a sparse variant that retains comparable performance with reduced computational cost~\cite{shen2025numerical}. Formally, given a set of model parameters \( \omega \), pruning aims to learn a binary mask \( m \in \{0, 1\}^{|\omega|} \) such that the effective parameters become \( \omega' = m \odot \omega \), where \( \odot \) denotes element-wise multiplication. The goal is to find \( m \) that minimizes the loss \( \mathcal{L} \) on task data while promoting sparsity:
\begin{equation}
\vspace{-2px}
\min_{m} F(m \odot \omega) + \lambda \|m\|_0,
\end{equation}

\noindent where \( \lambda \) controls the trade-off between task performance and sparsity. \textcolor{black}{Figure~\ref{fig:pruning_compression} (a) illustrates how model pruning reduces the size and complexity of neural networks, thereby facilitating their deployment on resource-constrained edge devices.}

In addition, pruning helps reduce both local computation and uplink communication by transmitting only the pruned subset of model parameters. Moreover, structured pruning (e.g., filter-level or layer-level) is often preferred for better compatibility with edge hardware. Adaptive pruning strategies have also been explored to cater to client heterogeneity, where each client prunes the model according to its own computational constraints and data characteristics. This flexibility makes pruning particularly well-suited for FL scenarios involving FMs, where devices often operate under diverse memory and statistical constraints. Numerous studies have investigated model pruning techniques to enable efficient FL at the network edge~\cite{jiang2022fedmp,jiang2023complement,yi2024fedpe,yu2025pruning}. FMs, such as large transformers, contain multiple layers, attention heads, and feed-forward modules that are often overparameterized for downstream tasks. \textcolor{black}{In FL with FMs, pruning enables each client to train or fine-tune a reduced version of the global model, significantly lowering computational costs and allowing resource-constrained devices to participate~\cite{bai2024fedspallm,fan2025ppc}. FedSpaLLM~\cite{bai2024fedspallm} is the first work to apply pruning techniques to LLMs within an FL framework, aiming to improve communication efficiency and mitigate system heterogeneity. It introduces a novel aggregation function that preserves important parameters by averaging only the non-zero elements across client models, thereby maintaining both sparsity and relevance. To further address limited client resources and the need to protect domain-specific knowledge in FL, PPC-GPT~\cite{fan2025ppc} compresses LLMs into task-specific small language models and employs a chain-of-thought distillation strategy to enable lightweight and effective local adaptation.} By enabling lightweight model versions that retain essential performance, model pruning plays a central role in making FL practical for large-scale foundation models, particularly in cross-device environments.  

\textbf{Model Compression.} Complementing pruning techniques, model compression strategies focus on minimizing communication cost in FFMs by reducing the size of model updates and memory usage without altering the model structure~\cite{zhu2024survey}. One widely adopted compression strategy is quantization, which reduces the precision of model parameters and gradients during transmission. By representing weights in lower bit-width formats (e.g., 8-bit or even binary), quantization dramatically reduces the size of model updates while still preserving most of the learning capacity. Formally, quantization maps the model parameter vector \( \omega \) into a discrete set of values \( \tilde{\omega} \), which can be expressed as
\begin{equation}
\vspace{-2px}
\min_{\tilde{\omega}} F(Q(\omega)) + \lambda \| \omega - Q(\omega) \|_2^2, \quad \tilde{\omega} = Q(\omega),
\end{equation}
where \( Q(\cdot) \) is the quantization function that projects the continuous values of \( \omega \) into a discrete space with lower bit precision. For instance, in the case of 8-bit quantization, the function \( Q(\omega) \) maps each parameter to one of 256 possible values. The objective is to minimize the loss \( F \) while controlling the quantization error, where \( \lambda \) balances the trade-off between model performance and compression error. This approach facilitates efficient communication, especially in FL scenarios with limited bandwidth or resource-constrained devices. \textcolor{black}{Figure~\ref{fig:pruning_compression} (b) illustrates the process of model compression through quantization, highlighting its impact on reducing model update size and transmission cost.}

\begin{table*}[t]
\centering
\textcolor{black}{\caption{\textcolor{black}{Taxonomy of Key Communication- and Computation-Efficient Strategies for Personalized Federated Intelligence.}}
\label{tab:CC_efficient_PFI}
\vspace{-10px}
\resizebox{0.99\linewidth}{!}{%
\renewcommand{\arraystretch}{1.3}
\begin{tabular}{@{}lll@{}}
\toprule
\textbf{Methods} & \textbf{Categories} & \textbf{Related Works} \\
\midrule
\multirow{3}{*}{Model Pruning} 
& Structured Pruning (\text{Comm. $\downarrow$} $\approx$ 80\%) & FedMP~\cite{jiang2022fedmp}, Complementary Pruning~\cite{jiang2023complement}, FedSpaLLM~\cite{bai2024fedspallm} \\
\cline{2-3}
& Adaptive Pruning  (\text{Comm. $\downarrow$} $\approx$ 90\%)& FedPE~\cite{yi2024fedpe}, PPC-GPT~\cite{fan2025ppc} \\
\cline{2-3}
& Unstructured Pruning  (\text{Comm. $\downarrow$} $\approx$ 70\%-90\%) & Sparse Pruning~\cite{yu2025pruning}, Numerical Pruning~\cite{shen2025numerical} \\
\midrule
\multirow{3}{*}{Model Compression} 
& Quantization-based Compression & CE-FFT (13\% Comm.)~\cite{zhang2025fft}, Quantized LoRA~\cite{jianhao2024promoting}, FedQLoRA (35\% Comp.)~\cite{hufedqlora} \\
\cline{2-3}
& Knowledge Distillation & FedID~\cite{ma2023fedid}, FedDAT(0.75\% Commu.)~\cite{chen2024feddat}, FedMKT(0.12\% Comp.) ~\cite{fan2024fedmkt} \\
\cline{2-3}
& Cluster-aware Distillation & FedCKMS (25-75\% Comp.)~\cite{wang2025cluster} \\
\bottomrule
\end{tabular}}}
\vspace{-15px}
\end{table*}

In FFMs, this is particularly valuable because even partial updates of large-scale models can be prohibitively significant in their raw form. CE-FFT~\cite{zhang2025fft} is one of the first works to investigate transmission efficiency in FFMs by proposing a per-channel quantization method that leverages the structural patterns of activations and parameters within LoRA modules. \textcolor{black}{Note that LoRA is a parameter-efficient fine-tuning technique that injects low-rank adaptation modules into pre-trained models, as introduced in Section 3.2.} Similarly, the authors in~\cite{jianhao2024promoting} integrate quantization strategies with LoRA to enable efficient parameter updates while significantly reducing communication overhead. To address quantization-induced bias, FedQLoRA~\cite{hufedqlora} introduces a quantization-aware adapter mechanism that compensates for quantization errors and further reduces memory usage during inference. These advances highlight quantization as a promising direction for enabling scalable, efficient communication in FFM systems without compromising model performance. Another effective approach is knowledge distillation, which serves as a powerful framework for model compression in FFMs~\cite{ma2023fedid,chen2024feddat,fan2024fedmkt,wang2025cluster}. This allows clients to benefit from the knowledge of the global model without storing or training the complete foundation model locally. FedID~\cite{ma2023fedid} first investigates the application of knowledge distillation to FL in the context of LLMs to mitigate the problem of misleading privileged knowledge caused by confirmation bias in previous works. By transferring the knowledge on an unlabeled public dataset, the proposed method reduces the communication overhead while achieving effectiveness in NLP tasks. Regarding knowledge transfer method, the authors in~\cite{kang2025grounding} propose the knowledge transfer-based FFMs framework that formulates problems of grounding FMs in FL, including corresponding objectives, representative knowledge transfer approaches, and privacy measurements. Recently, FedCKMS~\cite{wang2025cluster} introduces the cluster-based FL framework designed for efficient adaptation and fine-tuning of FMs across highly heterogeneous and resource-constrained edge devices. It combines partial training with cluster-aware knowledge distillation, enabling effective cross-cluster knowledge transfer from diverse sub-models to the central foundation model while significantly reducing communication and computation overhead. Overall, knowledge distillation in FFMs effectively transfers knowledge from large FMs to lightweight local models, enabling efficient personalization and training under strict resource and privacy constraints. \textcolor{black}{The key enabling communication- and computation-efficient technologies are summarized in Table~\ref{tab:CC_efficient_PFI}.} \textcolor{black}{Note that, in addition to model pruning and compression, meta-learning techniques such as MAML~\cite{finn2017model} can also be integrated with these strategies. This integration allows each client to not only adapt rapidly but also communicate and compute more efficiently with a compact model, thereby accelerating convergence and improving the scalability of personalized federated adaptation~\cite{fallah2020personalized_nips}.}

\subsection{Efficient Adaptation Strategies}

The scale of FMs presents a major barrier to their deployment in FL, where the communication bandwidth and device resources are severely constrained, especially on edge devices~\cite{sharshar2025vision}. Full fine-tuning of FMs on each client is impractical due to the model size and communication constraints. To address this, a class of techniques known as PEFT has emerged that enables client-side model adaptation by updating only a small fraction of the parameters~\cite{han2024parameter}. PEFT approaches are highly attractive in FL as they significantly reduce resource demands while allowing client-specific specialization.

\textbf{LoRA.} One of the most widely used PEFT strategies is LoRA~\cite{hu2022lora}, which introduces trainable low-rank matrices into the attention layer of the transformer model. \textcolor{black}{Instead of updating the full weight matrix, LoRA enables personalization in federated settings by allowing each client to learn its own low-dimensional adaptation while keeping the shared backbone frozen. Formally, for a pre-trained  weight matrix $\omega \in \mathbb{R}^{d \times k}$, each client learns a personalized increment $\Delta \omega$ defined as}
\begin{equation}
\vspace{-2px}
  \omega' = \omega + \Delta \omega = \omega + BA,
\end{equation}
\noindent where \( A \in \mathbb{R}^{d \times l} \) and \( B \in \mathbb{R}^{l \times k} \) are trainable client-specific low-rank matrices with \( l \ll \min(d, k) \), significantly reducing the number of trainable parameters. \textcolor{black}{This low-rank decomposition explicitly characterizes the personalization mechanism: each client obtains a distinct adaptation $BA$ that reflects its local data distribution, while all clients share the same frozen global weight $\omega$.} Typically, \( A \) is initialized from a normal distribution (e.g., \( A \sim \mathcal{N}(0, \sigma^2) \)) and \( B \) is initialized as zero, ensuring that the initial output remains unchanged. \textcolor{black}{The working mechanism of LoRA is illustrated in Appendix C.}

LoRA is particularly well-suited for FL due to its modular structure and minimal communication overhead. Clients can locally fine-tune only the low-rank matrices while keeping the backbone of the foundation model frozen. During communication rounds, only these compact matrices are transmitted to the server, substantially reducing bandwidth usage. Several studies~\cite{sun2024improving,wang2024flora,zhang2024towards_fedit} have explored LoRA for federated fine-tuning approaches. \textcolor{black}{FedIT~\cite{zhang2024towards_fedit} was the first work to apply LoRA in federated fine-tuning, demonstrating that transmitting low-rank updates can significantly reduce the communication cost. However, FedIT assumes that all clients use LoRA modules with same rank, which limits its practical applicability. In the practical systems, clients have heterogeneous computational budgets and therefore require different LoRA ranks to balance the personalization capacity. Due to this characteristic, mismatched shapes across LoRA matrices make the aggregation infeasible, preventing flexible deployment across devices. To address this limitation, subsequent works have proposed several strategies to overcome heterogeneous client ranks and aggregation inconsistency.} For example, FLora~\cite{wang2024flora} introduces an aggregation-noise-free federated fine-tuning approach that supports heterogeneous LoRAs by employing a stacking mechanism for aggregation. FFA-LoRA~\cite{sun2024improving} tackles data heterogeneity in federated fine-tuning by keeping the pre-trained LLMs frozen and initializing the non-zero components of the LoRA modules randomly, thus enhancing the privacy under differential privacy and improving computational efficiency. \textcolor{black}{Beyond the early LoRA-based federated works, several studies refine how LoRA updates are combined and initialized across heterogeneous clients. The authors in~\cite{guo2025selective} propose selective aggregation of LoRA parameters, updating only layers or ranks whose client updates exhibit high cross-client consistency-thereby reducing drift from non-IID data and improving stability under partial participation. LoRA-FAIR~\cite{bian2025lora} complements this idea with an aggregation-and-initialization refinement scheme that tackles the challenge of server-side aggregation bias in LoRA matrices and client-side initialization lag. These methods make LoRA aggregation more robust to statistical heterogeneity while preserving PEFT's communication advantages. }

\textbf{Prompt Tuning.} Another effective PEFT method is \textit{prompt tuning}, which adapts a foundation model by appending a small set of learned tokens, known as soft prompts, to the input. This method avoids changing internal model weights and steers the model towards downstream tasks using trainable embeddings~\cite{min2023recent}. Formally, given an input sequence \( x \), prompt tuning prepends a learnable prompt vector \( P \in \mathbb{R}^{l \times d} \), where \( l \) is the number of soft tokens and \( d \) is the model’s hidden size. The new input to the model becomes:
\begin{equation}
\vspace{-2px}
       F(\hat{x}); \quad \hat{x} = [P; x],
       \vspace{-2px}
\end{equation}

\noindent where only the parameters in \( P \) are updated during training, while the backbone model  \( F(\cdot) \)  remains frozen. This design enables lightweight and task-specific adaptation with minimal overhead. \textcolor{black}{The core working mechanism of prompt tuning is illustrated in Appendix D.}

Therefore, the lightweight nature of prompt vectors makes prompt tuning highly memory-and communication-efficient, well-suited to the constraints of FL. In FL scenarios, prompt tuning enables each client to maintain its personalized prompt vectors while sharing a globally frozen foundation model. This setup not only reduces the communication burden, as only the tiny prompt vectors are exchanged, but also offers better privacy protection since the majority of the model remains untouched and is less likely to reveal sensitive local data. Several recent studies have explored the adaptation of prompt tuning to FL, addressing key challenges such as domain generalization and data heterogeneity, as well as application areas including vision-language tasks and weather forecasting~\cite{zhao2023fedprompt,guo2023pfedprompt,li2024global}. FedPrompt~\cite{zhao2023fedprompt} was among the first to introduce prompt learning into FL, aiming to accelerate global aggregation and address scenarios with limited user data. To address statistical heterogeneity in a personalized manner, pFedPrompt~\cite{guo2023pfedprompt} introduces a client-specific attention module that generates locally tailored spatial visual features. To enable efficient model personalization, the authors in~\cite{li2024global} deploy the unbalanced optimal transport to align local visual features with corresponding local prompts, effectively balancing global consensus and local personalization. \textcolor{black}{Recently, the authors in~\cite{luo2025mixture} introduce a personalized MoE prompt mechanism in federated settings. Clients learn lightweight expert prompts and a routing policy that selects or blends experts per input/client distribution, improving transfer across skewed modalities while keeping communication small (only prompt and router parameters).} \textcolor{black}{In addition to soft prompts for prompt tuning, recent studies have also explored textual prompts in the context of foundation models. AutoPrompt~\cite{shin2020autoprompt} introduces a gradient-guided method for automatically generating discrete textual prompts, enabling language models to perform sentiment analysis. Notably, on the LAMA fact-retrieval benchmark, AutoPrompt achieves 43.3\% top-1 precision, substantially outperforming the single-prompt baseline of 34.1\%. Meanwhile, PromptSource~\cite{bach2022promptsource} offers a collaborative environment for designing, sharing, and evaluating human-crafted textual prompts across diverse tasks, with more than 2,000 prompts covering around 170 datasets.} Overall, prompt tuning offers a highly adaptable and communication-efficient mechanism for deploying FMs in FL. Its lightweight nature, modularity, and compatibility with privacy-preserving constraints make it a powerful tool for enabling PFI at scale.

\textbf{Adapter Tuning.} Beyond LoRA and prompt tuning, \textit{adapter tuning} has also gained attention for enabling efficient and modular fine-tuning of FMs in FL~\cite{cai2023efficient,chen2024feddat,long2024dual,saha2025fedpia}. Adapter tuning represents another line of PEFT methods, where small bottleneck modules are inserted between the layers of a pre-trained transformer~\cite{houlsby2019parameter}. Specifically, an adapter module typically consists of a down-projection, a nonlinearity, and an up-projection, formulated as follows:
\begin{equation}
\vspace{-2px}
    \text{Adapter}(h) = h + \omega_\text{up} \, f(\omega_\text{down} \, h),
\end{equation}

\noindent where \( h \) is the hidden activation from the Transformer layer, \( \omega_\text{down} \in \mathbb{R}^{m \times d} \) projects the features into a low-dimensional space (\( m \ll d \)), \( f(\cdot) \) is a non-linear activation function (e.g., ReLU), and \( \omega_\text{up} \in \mathbb{R}^{d \times m} \) projects back to the original dimension. The residual connection ensures that the pre-trained model's behavior is preserved when the adapter is initialized to near-zero impact. Only the adapter parameters are updated during fine-tuning, making the approach both parameter-efficient and modular. We illustrate the key design of the adapter architecture in Appendix E.

\begin{table*}[t]
\centering
\textcolor{black}{\caption{\textcolor{black}{Taxonomy of Key Adaptation Strategies for Personalized Federated Intelligence.}}
\label{tab:adaption_efficient_PFI}
\vspace{-10px}
\resizebox{0.99\linewidth}{!}{%
\renewcommand{\arraystretch}{1.3}
\begin{tabular}{@{}lll@{}}
\toprule
\textbf{Methods} & \textbf{Categories} & \textbf{Related Works} \\
\midrule
\multirow{2}{*}{LoRA} 
& Standard LoRA & Hu \textit{et al.} (0.025\% trainable parameters)~\cite{hu2022lora} \\
\cline{2-3}
& Federated LoRA & FedIT (0.26\% params)~\cite{zhang2024towards_fedit}, FLora (21.5\% params)~\cite{wang2024flora}, FFA-LoRA~\cite{sun2024improving}, Guo \textit{et al.}~\cite{guo2025selective}, LoRA-FAIR~\cite{bian2025lora} \\
\midrule
\multirow{2}{*}{Prompt Tuning} 
& Soft Prompts & Min \textit{et al.}~\cite{min2023recent},  FedPrompt (0.007-0.014\% Comm.)~\cite{zhao2023fedprompt}, pFedPrompt~\cite{guo2023pfedprompt}, Li \textit{et al.}~\cite{li2024global} \\
\cline{2-3}
& Textual (Hard) prompts & AutoPrompt~\cite{shin2020autoprompt}, PromptSource~\cite{bach2022promptsource} \\
\midrule
\multirow{2}{*}{Adapter Tuning} 
& Single Adapter & Houlsby \textit{et al.}~ (1.14\% params)\cite{houlsby2019parameter}, FedAdapter (0.48-0.55\% params)~\cite{cai2023efficient} \\
\cline{2-3}
& Dual Adapter & FedDAT (0.75\% Commu.)~\cite{chen2024feddat}, FedDPA (0.06\% Commu.)~\cite{long2024dual}, FedPIA~\cite{saha2025fedpia} \\
\bottomrule
\end{tabular}}}
\vspace{-15px}
\end{table*}

Therefore, this design is especially attractive in FL, where devices operate under limited resources and direct access to global model parameters is often restricted. In federated settings, adapter tuning enables clients to personalize the model by training only their local adapter modules while sharing or aggregating them selectively to improve generalization. \textcolor{black}{To reduce NLP training costs, FedAdapter~\cite{cai2023efficient} employs a progressive training paradigm with trial-and-error profiling while training only adapter modules, enabling convergence within just a few hours-up to 155.5× faster than vanilla FedNLP and 48× faster than strong baselines.} In the domain of vision-language tasks, FedDAT~\cite{chen2024feddat} is the first to propose a fine-tuning framework for heterogeneous federated settings by leveraging a dual-adapter teacher model. Their approach regularizes client updates and employs mutual knowledge distillation to enable efficient knowledge transfer. \textcolor{black}{Moreover, FedDPA~\cite{long2024dual} introduces a dual-personalization design that separates server-shared and client-specific adapter components, enabling global transfer while preserving local specialization. Recently, FedPIA~\cite{saha2025fedpia} enhances the integration of FL and PEFT by enabling richer information exchange between server–client adapters and local–global adapters within each client, achieving F1-score improvements of 5.1\% on Open-I and 2.73\% on MIMIC over FedDAT across all clients and adapter configurations.} Specifically, the method permutes client adapter neurons at each layer to align with globally initialized adapter neurons, guided by the theory of Wasserstein Barycenters, thereby improving cross-client consistency and personalization. Overall, adapter tuning offers a strong balance between model adaptability and communication efficiency, making it a practical choice for real-world FL systems involving foundation models across diverse user contexts and devices. \textcolor{black}{We summarize the key adaptation technologies for enabling personalized federated intelligence in Table~\ref{tab:adaption_efficient_PFI}. }

\textcolor{black}{\textbf{Trade-offs among PEFT techniques:} Although PEFT methods such as LoRA, prompt tuning, and adapter tuning all aim to enable lightweight personalization of FMs, they exhibit distinct trade-offs in practical personalized federated intelligence deployments. Prompt tuning achieves the lowest communication and memory overhead by only optimizing a small set of input embeddings. However, its limited adaptation capacity may lead to suboptimal performance under severe statistical heterogeneity or complex domain shifts. Adapter-based methods introduce dedicated trainable modules within the model, providing stronger personalization robustness and stability across heterogeneous clients, at the cost of increased memory footprint and aggregation complexity. LoRA offers a flexible middle ground, where the choice of rank directly governs the balance between personalization expressiveness and efficiency: low-rank settings reduce communication overhead and overfitting risk, while higher-rank configurations improve adaptation accuracy but may amplify personalization drift in highly non-IID settings. Overall, choosing among PEFT strategies requires jointly considering deployment constraints, personalization granularity, and the degree of cross-client heterogeneity to achieve an optimal balance between efficiency and performance.}

\subsection{Summary and Lessons Learned}
This section reviewed two complementary technique families, model compression and PEFT, that are essential for enabling efficient deployment of FMs. Both approaches mitigate the communication and computation constraints of edge devices while preserving personalization capability.

\textbf{Takeaways:}
Model compression methods (e.g., pruning, quantization, and knowledge distillation) reduce model size and inference cost, facilitating efficient transmission and local execution. PEFT techniques, such as LoRA, prompt tuning, and adapter tuning, enable lightweight personalization by updating only a small subset of parameters. Together, these methods provide flexible and scalable adaptation of large models to heterogeneous clients.

\textbf{Connection to PFI:}
Compression and PEFT are key enablers of PFI, as they significantly reduce the resource footprint of FMs and allow diverse clients to participate in federated training and inference while supporting privacy-preserving personalization. Open challenges include performance degradation under aggressive compression, limited expressiveness of PEFT under extreme client heterogeneity, dependency on proxy data for distillation, and difficulties in aggregating adapter- or LoRA-based updates. Dynamically selecting client-specific strategies in decentralized settings remains an important direction for future research.

\section{Towards Trustworthy Personalized Federated Intelligence}
\label{sec_trustworthy_PFI}

\textcolor{black}{Building upon efficient personalization, the next stage of PFI focuses on ensuring that adapted models remain trustworthy in real-world deployment. This section examines the evolving landscape of trustworthy PFI, beginning with fairness considerations and progressing through interpretability, hallucination mitigation, robustness against various attacks, and privacy protection, thereby establishing a reliable foundation for subsequent adaptive-driven refinement.}

\subsection{Mitigating Bias and Ensuring Fairness}
Fairness in FMs refers to the model's ability to avoid discriminating against individuals or groups within society~\cite{kaur2022trustworthy_survey}. Since FMs are often trained on diverse datasets collected from numerous clients, they are susceptible to inheriting and amplifying biases present in the data, which can result in unfair or discriminatory outcomes. To mitigate such issues, we explore three complementary strategies to enhance fairness in PFI: model-level processing, client-side processing, and server-side processing.

\textbf{Model-level Processing.}
Given the privacy constraints inherent to FMs, a complementary strategy to mitigate bias is to modify the FM architecture or adapt the training objective to include fairness-aware components. For example, a practical and widely adopted approach involves decoupling the model into shared base layers and a personalized head, such as FedPer~\cite{arivazhagan2019federated}, enabling collaborative learning of generalizable base layers while allowing client-specific adaptations.

\textbf{Client-side Processing.}
In addition to model-level processing, client-side data preprocessing has been explored as a means for bias mitigation in centralized settings~\cite{sablayrolles2020radioactive}. However, in FL, individual user data remains inaccessible to other parties, including servers. An alternative approach to tackle this challenge is integrating democratized learning~\cite{minh2021demai}, which can leverage client-level model characteristics to enhance fairness while preserving personalization. By exploiting similarities across clients’ model updates or representations, it becomes possible to promote fairness among participating clients without compromising local preferences. \textcolor{black}{A complementary line of work adopts proactive bias-aware aggregation, such as FedFair~\cite{chu2021fedfair}, which evaluates each client’s contribution based on both bias metrics and personalized objectives before global aggregation. Clients with excessive bias can then be downweighted or excluded to prevent unfair influence on the shared model. However, such approaches may remain sensitive to non-IID data, where heterogeneity may cause certain clients to be misclassified as biased despite the absence of systematic unfairness~\cite{sabah2025fairdpfl}. More recent studies, such as FairDPFL-SCS~\cite{sabah2025fairdpfl}, explicitly highlight this limitation and introduce dynamic client-selection strategies that balance fairness, accuracy, and personalized utility under non-IID data, yielding dramatic accuracy gains from 58.21\% to 99.04\% and thereby demonstrating the clear benefits of aggregating local updates.}

\textbf{Server-side Processing.} Another important approach to mitigating bias involves applying client-wise evaluation and penalization, post-aggregation correction, and fair routing in MoE architectures on the server side to adjust model behaviors or representations and promote fairness. By replacing unfair word embeddings within the models, the model bias can be effectively reduced, thereby enhancing fairness~\cite{bolukbasi2016debiasing_embedding}. One notable approach is adjusting the output of FMs without altering their parameters, which involves leveraging the features learned by FMs to modify the model output according to desired specifications. This innovative technique, as demonstrated in~\cite{li2024self}, involves interpreting the meaning of feature vectors and adapting model outputs accordingly to address bias and ensure safety content in the system. 
Overall, the coordinated examination of key enabling technologies across the client, model, and server sides is expected to collectively benefit PFI, resulting in reduced bias and enhanced fairness.

\subsection{Interpretable Capabilities}

Interpretable capabilities are key to building trust in PFI, where models operate on sensitive, user-specific data. To this end, we focus on two complementary approaches: traditional XAI techniques and emerging reasoning-based methods. XAI enhances transparency by revealing how models arrive at decisions, either at a global model level or for individual predictions. Meanwhile, reasoning models improve interpretability by generating structured intermediate steps, aligning model behavior with human-like logic. Together, these two selected techniques provide clearer insights into personalized model behavior, making the decision in PFI more transparent and trustworthy.

\textbf{XAI.} Various strategies can be proposed to enhance model performance in the context of FL within FMs~\cite{tan2022towards}, but these often result in opaque models with decision-making processes that are difficult to interpret. This lack of transparency can undermine user trust and accountability~\cite{lopez2024interplay}. XAI techniques offer a solution by providing insights into how models arrive at their predictions, helping to bridge this transparency gap. Based on the scope of explanation, XAI methods can be broadly categorized into two types: local explanations, which focus on individual predictions, and global explanations, which aim to interpret the model's overall behavior. \textcolor{black}{The core principles and workflows of both local and global explainability approaches are illustrated in Appendix F.}

Popular methods such as GradCAM~\cite{selvaraju2017grad}, AttentionViz~\cite{yeh2024attention_viz}, and SHAP~\cite{lundberg2017unified} provide local explanations by illustrating how specific features influence individual predictions. For instance, GradCAM highlights the regions of the input that had the most influence on the model's decisions, while AttentionViz leverages attention mechanisms to visualize how input features are weighted during processing. Additionally, SHAP enhances interpretability by assigning Shapley values to input features, quantifying their contributions to the predictions based on game-theoretic principles. Building on this, FedSA~\cite{otmani2024fedsv} adapts SHAP to federated settings by estimating each client's contribution, thereby enabling the detection of and defense against malicious participants in the federation. In contrast, global explanations, such as model distillation~\cite{tan2018distill}, GAM~\cite{caruana2015intelligible}, and LIME~\cite{ribeiro2016should}, offer a broader understanding of the model's decision-making processes across all data, helping users grasp how the model operates as a whole. For instance, LIME~\cite{ribeiro2016should} treats the model as a black box and approximates its local decision boundary by fitting an interpretable surrogate model on perturbed samples around a specific input. Formally, LIME solves the following optimization problem:
\begin{equation}
\vspace{-2px}
\arg\min_{g \in G} \ F(f, g, \pi_x) + \Omega(g),
\vspace{-3px}
\end{equation}
where \( f \) is the original black-box model, \( g \in G \) is an interpretable surrogate model (e.g., linear regression), \( F(f, g, \pi_x) \) measures the fidelity of \( g \) in approximating \( f \) near the input \( x \) using a locality-aware weighting function \( \pi_x \), and \( \Omega(g) \) is a regularization term that penalizes model complexity to ensure interoperability. Additionally, the distill-and-compare approach proposed by~\cite{tan2018distill} allows for auditing black-box risk scoring models by distilling them into transparent student models and comparing these to ground-truth models, thereby offering valuable insights into the black-box model's decision-making process. Therefore, integrating these XAI methods into the development of the PFI process shows great promise in ensuring that the model’s decision-making process remains transparent and trustworthy to users.

\textbf{Reasoning Methods.}
FMs can enhance their reasoning capabilities through the integration of various techniques, including chain-of-thought (CoT)~\cite{fan2025ppc}, self-consistency with CoT (CoT-SC)~\cite{wang2022self}, and tree-of-thought (ToT)~\cite{yao2023tree}. The difference between them is as shown in Appendix G. CoT facilitates a systematic approach to problem-solving, guiding the model through logical steps rather than directly jumping to conclusions. This method enhances the model's proficiency in handling complex tasks requiring multiple steps, such as mathematical problems, logic puzzles, and decision-making processes. By incorporating CoT, FMs are encouraged to provide intermediate steps to justify their answers, thereby offering more insight into the reasoning behind the model's outputs. Building upon CoT, CoT-SC was introduced to enhance the reasoning process by exploring multiple paths simultaneously. In CoT-SC, several CoTs are generated during the inference phase of FMs, allowing them to navigate through various logical steps towards the correct solution. These CoTs are generated independently, and the final output is determined based on the most frequently occurring solution among the CoTs. On the other hand, ToT takes this process a step further by fostering connections among these independent CoT-SC pathways. In ToT, solutions are not just evaluated independently; they also share explanations, collaboratively refining reasoning to formulate the best possible answer. By aligning different perspectives, ToT enables FMs to generate more comprehensive explanations and identify the optimal decision for a given problem.

In the context of applying reasoning abilities within PFI, we utilize RL with feedback to enhance the reasoning capabilities of FMs. However, privacy concerns present challenges in implementing techniques such as RLHF~\cite{ouyang2022training} and RLAIF~\cite{bai2022constitutional} within PFI. A more detailed discussion of these issues will follow in the next section. Conversely, the pure RL approach~\cite{guo2025deepseek} has demonstrated promising reasoning potential for FMs. Unlike methods that require training a separate reward model for integration into PFI, pure RL directly generates rewards using predefined rules. Furthermore, integrating pure RL into PFI enables a more adaptive learning process. By leveraging predefined reward criteria, FMs can refine their reasoning without relying on external validation or manually labeled data. This independence makes pure RL particularly suitable for PFI, where privacy and decentralization are paramount considerations.

\vspace{-0.1in}

\subsection{Mitigating the Challenges of Hallucinations} 
Although FMs can already provide strong generalization capabilities, they still tend to produce false or fabricated details, also known as hallucinations. Such hallucination reduces the reliability of FMs by generating content that is nonsensical or unfaithful to the source material~\cite{maynez2020faithfulness}. Based on the alignment failures between model outputs and either real-world facts or the provided inputs, we discuss two types of hallucination, including factuality hallucination and faithfulness hallucination~\cite{huang2023survey}. 

Factuality hallucinations arise when FMs produce outputs that contradict real-world facts or provide unverifiable claims. These can be further divided into two subtypes: factual contradictions, which involve inaccuracies about entities or relationships (e.g., misattributing inventions), and factual fabrications, which consist of unverifiable statements or exaggerated claims. In contrast, faithfulness hallucinations occur when FMs fail to align with user instructions, provided context, or logical reasoning. These deviations are categorized into three subtypes: instruction inconsistency, where outputs unintentionally misalign with directives; context inconsistency, where responses contradict user-provided contextual details; and logical inconsistency, which involves contradictions in reasoning or discrepancies between intermediate steps and final conclusions. To mitigate the challenges of hallucinations in the development of PFI, various strategies are typically employed across data preparation, training methodologies, and inference processes~\cite{huang2023survey}. For instance, in data preparation, techniques such as data augmentation and the careful curation of high-quality training data can help reduce the occurrence of hallucinations~\cite{huang2023survey}. During training, methods like RLHF~\cite{ouyang2022training} and RLAIF~\cite{bai2022constitutional} can help align model outputs more closely with human expectations and real-world facts. In the inference phase, approaches such as RAG~\cite{izacard2023atlas} and contrastive decoding~\cite{wang2024mitigating} can ensure that the generated content remains both factual and contextually relevant. However, within the constraints of the PFI scenario, the lack of access to client data limits the feasibility of approaches reliant on data. For instance, applying data filtering~\cite{huang2023survey} to ensure high-quality and accurate information requires clients to possess strong domain expertise. This is impractical, as data is collected from a diverse range of devices, making it impossible to guarantee its consistency or reliability. As a result, minimizing hallucinations during the training phase becomes a critical focus for improving the performance and reliability of FMs in PFI.

Effective techniques for improving reasoning in PFI include RLHF~\cite{ouyang2022training} and RLAIF~\cite{bai2022constitutional}. These methods help refine the reasoning capabilities of FMs by leveraging feedback-driven optimization. For an in-depth discussion and comparative analysis of RLHF and RLAIF, readers are referred to~\cite{qiao2025deepseek}. However, implementing RLHF in PFI presents significant privacy challenges, as it typically requires sampling and training a reward model using client datasets~\cite{qiao2025deepseek}. \textcolor{black}{A possible solution for this challenge is collaborative training that can effectively optimize both the FM and the reward model. In this approach, each client independently generates reasoning data from their local dataset and trains a reward model alongside the FM.} Given the large number of clients, the aggregated reward model reflects the collective intent of the broader community, ensuring a more balanced and generalized training process. \textcolor{black}{However, one limitation of this approach is the requirement for additional labeled data, which can be expensive and resource-intensive. In real-world scenarios, obtaining consistent labeled data across multiple clients poses significant challenges due to data heterogeneity and privacy constraints. A scalable alternative, such as RLAIF~\cite{bai2022constitutional}, can help reduce reliance on manual annotation during rewording model training. However, this technique often leverages off-the-shelf LLMs, which are prone to generating inaccurate or fabricated outputs, especially when fine-tuned on low-quality datasets. Therefore, further empirical studies are necessary to address and minimize hallucination risks in FMs within PFI.}

\subsection{\textcolor{black}{Enhancing Robustness Against Attacks and Privacy Protection}} 
\textcolor{black}{Beyond mitigating bias and improving interpretability, enhancing robustness against adversarial attacks is a critical component of FL. Such attacks can manipulate model updates, degrade performance, or alter behavior across FL frameworks, creating serious security concerns in real-world deployments. While safety is a fundamental baseline requirement for all FL systems, it is especially crucial in the context of PFI, where FMs demand higher accuracy and reliability given their broad and sensitive applications. Addressing robustness and safety challenges is therefore essential, as adversarial threats pose unique risks to the trustworthy personalization of FMs.} These attacks encompass adversarial examples~\cite{qiao2024logit}, where subtle input perturbations are used to mislead model predictions; backdoor attacks~\cite{gong2022backdoor}, in which malicious clients inject triggers to manipulate an FM’s behavior on specific inputs; model poisoning attacks~\cite{fang2024byzantine}, where adversarial participants deliberately corrupt FM updates during federated training; and inference attacks~\cite{hu2022membership}, in which adversaries attempt to extract sensitive information from shared model updates or predictions.

\subsubsection{\textcolor{black}{Adversarial Examples}}
\textcolor{black}{The increasing adoption of FMs in FL introduces new vulnerabilities, as their high-dimensional input space, large parameter size, and modular adaptation layers create a substantially larger attack surface compared to traditional FL models~\cite{hu2024federated}. This makes it critical to explore defense mechanisms tailored to the unique challenges of FL-FMs. To counter adversarial examples, FatCC~\cite{qiao2024logit} combines global feature contrastive learning with adversarial training, providing an effective defense mechanism in traditional FL settings. However, its effectiveness in the context of FMs remains uncertain, as the large scale and complexity of FMs can amplify vulnerabilities. Moreover, full-scale adversarial training becomes computationally prohibitive for FMs due to their size. To address these challenges, recent approaches have explored alternative strategies. For instance, FedEAT~\cite{pang2025fedeat} applies embedding-space adversarial training combined with robust aggregation (geometric median) to improve the adversarial robustness of federated LLMs with minimal performance loss. Other lightweight methods, such as pre-trained model-guided adversarial fine-tuning~\cite{wang2024pre} and adversarial adaptation techniques like AdvLoRA~\cite{ji2024advlora}, aim to enhance FM robustness efficiently without incurring the high cost of full-scale adversarial training. } 

\subsubsection{\textcolor{black}{Backdoor and Poisoning Attacks}}
\textcolor{black}{In addition to adversarial examples, FL-FMs are also susceptible to other malicious interventions, such as backdoor and poisoning attacks, which require dedicated defense mechanisms. To mitigate backdoor attacks, SDFC~\cite{huang2024fisher} employs Fisher calibration to identify and regulate distribution-sensitive parameters based on parameter consistency, thereby prioritizing clients whose updates align with the global distribution and effectively suppressing the influence of malicious participants in heterogeneous federated settings. Beyond parameter-level manipulations, recent studies such as~\cite{chaudhari2024phantom} also highlight the threat of backdoor injection in RAG, where attackers poison the retrieval corpus, for example, by inserting trigger-containing documents that cause the FM to output targeted responses whenever the trigger appears, thus enabling covert and persistent manipulation of downstream FM behaviors. Another line of defense leverages the modular structure of FL-FMs, recognizing that securing these models is substantially more challenging, particularly because attackers can exploit FM vulnerabilities to inject backdoors through synthetically generated data. To address this, the authors in \cite{bi2025securing} propose a data-free defense strategy that constrains internal activations to remain within safe, expected ranges, thereby suppressing backdoor behaviors while preserving the functionality of the FM. These activation constraints are optimized jointly with FL training using synthetic data, resulting in an effective and scalable defense mechanism tailored to the unique challenges of FL-FM settings. To defend against model poisoning attacks, \cite{yazdinejad2024robust} introduces an internal auditor that analyzes encrypted gradients using Gaussian mixture models and Mahalanobis distance, achieving robust aggregation with minimal computational and communication overhead.}

\subsubsection{\textcolor{black}{Inference Attacks}}
\textcolor{black}{Beyond active attacks on model behavior, inference attacks~\cite{hu2022membership} pose another serious threat to user privacy, as adversaries attempt to extract sensitive information from shared model updates or predictions. In traditional FL, studies such as deep leakage from gradients (DLG)~\cite{zhu2019deep} have shown that an honest-but-curious server can reconstruct users’ private training data by optimizing dummy inputs to match the shared gradients, revealing that gradient exchange alone can expose sensitive information. In the context of FL-FMs, recent work~\cite{sami2025gradient} has further demonstrated that attackers can perform gradient-inversion attacks on PEFT workflows, exploiting even lightweight adapter gradients to reconstruct high-fidelity local training samples. To counter such attacks, methods like FLSG~\cite{fan2023flsg} have been proposed, which defend against passive label inference attacks by generating and using Gaussian-distributed gradients while maintaining lower computational costs. Additionally, FedShield-LLM~\cite{mia2025fedshield} proposes a secure and efficient federated fine-tuning framework that combines fully homomorphic encryption (FHE)-protected and pruned LoRA updates to defend against inference attacks while significantly reducing computational and communication overhead for resource-constrained clients. Although these defenses do not fully address the underlying attack issues, their proposed mechanisms show potential for balancing robustness and efficiency, making them relatively suitable for deployment in personalized federated settings involving large-scale models. }

\subsubsection{\textcolor{black}{Security and Privacy}}

\textcolor{black}{At a broader level, the continued adoption of FL-FMs highlights the need for a unified security and privacy framework, such as confidential federated learning, that can jointly strengthen robustness and privacy across diverse threat vectors. In this direction, secure aggregation~\cite{so2022lightsecagg} ensures that individual client updates remain hidden from the server even when training high-dimensional FM adapters, such as LoRA-based updates in federated tuning of LLMs. Differential privacy~\cite{wei2020federated} further enhances protection by injecting carefully calibrated noise into updates, offering formal privacy guarantees against inference attacks in federated large models. Moreover, trusted execution environments~\cite{mo2021ppfl} provide hardware-backed isolation that safeguards both computations and sensitive parameters in FL-FM pipelines, protecting them from untrusted servers or malicious insiders and thereby enabling secure and privacy-preserving PFI. We summarize the key trustworthiness technologies for enabling personalized federated intelligence in Appendix H.}

\subsection{Summary and Lessons Learned}
This section has examined five core dimensions for enabling trustworthy PFI: bias mitigation, fairness, interpretability, hallucination mitigation, and robustness. Bias and fairness address disparities arising from pre-training and deployment through multi-level mitigation strategies. Interpretability enhances transparency via XAI and reasoning-based generation, while hallucination mitigation improves factuality using data filtering and feedback-driven optimization (e.g., RLHF and RLAIF). \textcolor{black}{Robustness is reinforced against adversarial, backdoor, poisoning, and inference attacks through lightweight and privacy-preserving defenses such as AdvLoRA and Fisher calibration. AdvLoRA enhances robustness to adversarial examples in FL-FMs, while Fisher calibration suppresses backdoor behaviors by identifying distribution-sensitive parameters. Although they defend against different classes of threats, both approaches align with and contribute to the overarching goal of trustworthy PFI.}

\textbf{Takeaways:} Bias, fairness, interpretability, hallucination mitigation, and robustness collectively enable equitable, transparent, factual, and secure personalization, forming the foundation of trustworthy PFI.

\textbf{Connection to PFI:} These dimensions are essential for trustworthy PFI, ensuring fair and accountable personalization, faithful generation, and resilience against adversarial, backdoor, and privacy-invasive attacks. However, open challenges include limited access to sensitive attributes for fairness correction, privacy–explainability trade-offs, supervision scarcity for hallucination mitigation in decentralized settings, and efficiency–robustness trade-offs.

\section{Towards Adaptive Personalized Federated Intelligence}
\label{sec_adaptive_PFI}

\textcolor{black}{Beyond efficient and trustworthy deployment, the final stage of PFI lies in continuous adaptation to evolving user contexts. While efficient and trustworthy mechanisms lay the foundation for deployment, static models may still struggle with outdated knowledge. To address this, integrating RAG offers a promising solution by enabling adaptive and context-aware generation. This section explores the role of RAG in PFI by examining its foundations and design space.}

\subsection{Fundamentals of Retrieval-Augmented Generation}

\begin{wrapfigure}{r}{0.45\textwidth}
    \centering
    \vspace{-10px}
    \includegraphics[width=\linewidth]{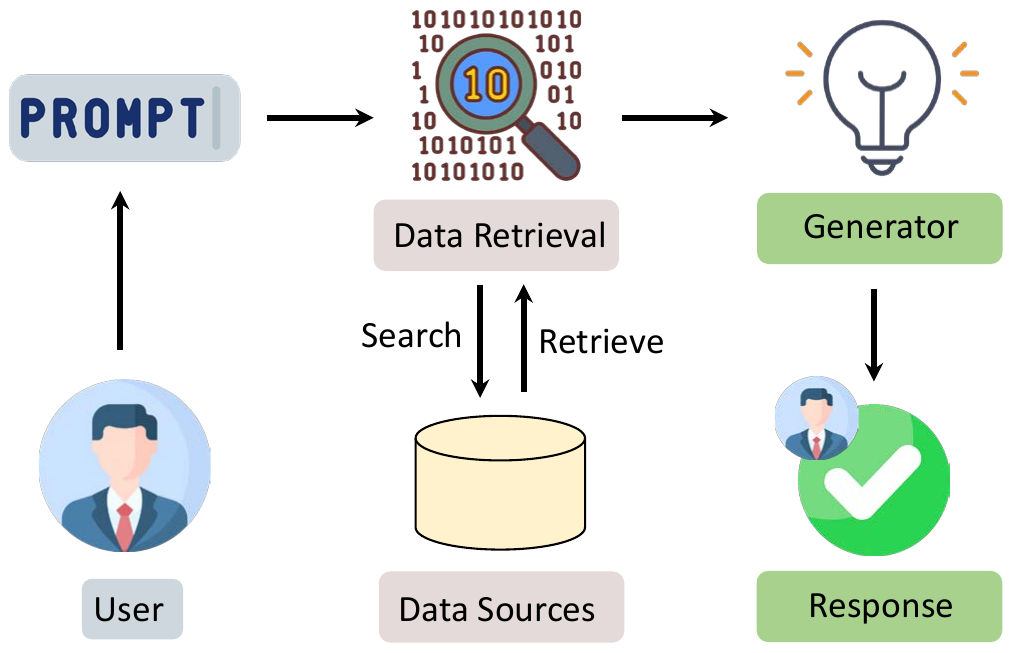}
    \caption{The working mechanism of RAG~\cite{educba_rag}.}
    \label{fig:RAG}
    \vspace{-10px}
\end{wrapfigure}

\textcolor{black}{RAG~\cite{lewis2020retrieval} is a hybrid framework that integrates neural retrieval with conditional generation or prediction by explicitly conditioning the model on retrieved external evidence items.} \textcolor{black}{Here, an \emph{evidence item} $d$ denotes a generic retrieved unit of external information (e.g., passage, structured record, table row, feature vector, or multimodal snippet) used to condition prediction or generation.}
\textcolor{black}{In the context of PFI, RAG can be understood as decomposing personalized intelligence into two cooperative functions: a retriever that identifies information relevant to the client’s current input, and a predictor/generator that produces the desired output using both the input and the retrieved evidence items.}
\textcolor{black}{Accordingly, we can represent the personalized model for client \(k\) as \(\omega_k=\{\omega_k^R,\omega_k^G\}\), where \(\omega_k^R\) parameterizes the retriever \(\mathcal{R}_{\omega_k^R}\) and \(\omega_k^G\) parameterizes the predictor/generator \(\mathcal{G}_{\omega_k^G}\).}

\textcolor{black}{Different from purely parametric models that rely solely on internal weights, RAG augments inference by retrieving relevant evidence items at inference time.
The retriever is typically implemented as a dense bi-encoder~\cite{karpukhin2020dense}, where both the input and the evidence items are encoded into a shared embedding space.
The retriever score between an input \(x\) and an evidence item \(d\) parameterized by \(\omega_k^R\) is computed as follows:
\begin{equation}
\vspace{-2px}
\mathcal{R}_{\omega_k^R}(x,d)=\text{sim}(E_x(x;\omega_k^R),E_d(d;\omega_k^R))=E_x(x;\omega_k^R)^\top E_d(d;\omega_k^R),
\end{equation}
where \(E_x(\cdot)\) and \(E_d(\cdot)\) denote the input and evidence item encoders, respectively, and \(\text{sim}(\cdot)\) represents dot product similarity.
The retriever selects the top-\(N\) evidence items \(\{d_1,d_2,\ldots,d_N\}\) with the highest scores for a given input.}
\textcolor{black}{To obtain a retrieval probability distribution, we normalize the scores over the retrieved set using a softmax function with temperature \(\tau\):
\begin{equation}
\vspace{-2px}
P(d_i|x;\omega_k^R)=\frac{\exp(\mathcal{R}_{\omega_k^R}(x,d_i)/\tau)}{\sum_{j=1}^{N}\exp(\mathcal{R}_{\omega_k^R}(x,d_j)/\tau)}.
\end{equation}}

\textcolor{black}{In practice, the top-$N$ items are retrieved, and the scores are renormalized over this set. Once the relevant evidence items are retrieved, the predictor/generator \(\mathcal{G}_{\omega_k^G}\) produces the final output \(y\). Depending on the task, \(\mathcal{G}_{\omega_k^G}\) may correspond to a conditional language model, a multimodal network, or a task-specific predictor that consumes \((x,d_i)\).
The output is conditioned on the original input \(x\) and the retrieved evidence items. To aggregate evidence from multiple retrieved evidence items, the likelihood of producing the output given the personalized model \(\omega_k\) is computed by marginalizing over the retrieved evidence items as follows:
\vspace{-2mm}
\begin{equation}
\vspace{-2px}
P(y|x;\omega_k)=\sum_{i=1}^{N} P(y|x,d_i;\omega_k^G)\,P(d_i|x;\omega_k^R),
\end{equation}}

\noindent \textcolor{black}{where \(P(y|x,d_i;\omega_k^G)\) denotes the probability of producing \(y\) given the input \(x\) and retrieved evidence item \(d_i\). The training objective of RAG is to maximize the likelihood of the correct task output. Within the PFI framework, this objective defines the local loss function \(F_k(\omega_k)\) for client \(k\), computed as the empirical risk over the client’s local dataset \(\mathcal{D}_k\):
\vspace{-2mm}
\begin{equation}
\vspace{-2px}
F_k(\omega_k)=\frac{1}{|\mathcal{D}_k|}\sum_{(x,y)\in\mathcal{D}_k}-\log\sum_{i=1}^{N} P(y|x,d_i;\omega_k^G)\,P(d_i|x;\omega_k^R).
\end{equation}}

\textcolor{black}{The retrieval probability \(P(d_i|x;\omega_k^R)\) serves as a mixture weight over retrieved evidence items, allowing the predictor/generator to synthesize evidence items into reliable and context-aware outputs. Note that in PFI, adaptation may occur either by optimizing (parts of) $\omega_k^R$ and/or $\omega_k^G$ via FL, or by keeping parameters fixed and updating the external knowledge/index used for retrieval. The working mechanism of RAG at inference time is shown in Fig.~\ref{fig:RAG}.}

\subsection{Design Space for Adaptive Personalized Federated Intelligence}
Building on the foundational principles of RAG outlined above, this subsection investigates how retrieval-based mechanisms can be effectively adapted to support user-centric personalization in PFI. Below, we structure the design space into two complementary dimensions: client-side personalization and server-side optimization.

\subsubsection{Client-side~Personalization} To enable personalization on the client side, various lightweight and privacy-preserving techniques such as prompt conditioning, knowledge indexing, and retriever adaptation can be developed.

\textbf{Prompt conditioning.}
\textcolor{black}{
Prompt conditioning is an inference-time mechanism for steering a \emph{frozen} FM by prepending or otherwise augmenting the input with a task- or user-specific prompt, rather than updating model parameters~\cite{brown2020language}.
When the prompt includes instructions and/or exemplars, the FM may exhibit \emph{in-context learning}, i.e., apparent task adaptation induced purely by the provided context without gradient-based optimization~\cite{brown2020language}.}

\textcolor{black}{
Formally, given a query \(x\), a client-specific prompt \(p_k\), and a frozen FM with parameters \(\omega\), generation is conditioned on the concatenated input:
\begin{equation}
\vspace{-2px}
    y \sim p_{\omega}\!\left(y \mid [p_k; x]\right), \qquad \omega \ \text{is fixed}.
\end{equation}
Here, \([p_k; x]\) denotes token-level concatenation of the prompt and query.
The prompt \(p_k\) may be a natural-language instruction, a set of exemplars (few-shot context), or a retrieved text prefix (e.g., policy, persona, or domain hints) prepended to the input \cite{brown2020language,lewis2020retrieval}.
Such conditioning can control task specification, output format, and domain emphasis without modifying the backbone parameters.}

\textcolor{black}{Prompt conditioning is particularly attractive in federated or multi-client settings, since client-specific behavior can be adjusted via client-side context selection rather than exchanging full-model updates. Recent \emph{federated in-context} frameworks explicitly exploit this property to reduce communication by sharing natural-language context instead of model parameters\cite{wu2024federated, wang2025federated}. Prompt conditioning also integrates naturally with RAG: retrieved evidence is injected into the input and thus serves as additional contextual conditioning for generation \cite{lewis2020retrieval}.
Moreover, prompts can shape retrieval intent (e.g., query rewriting/augmentation) and guide how retrieved evidence is interpreted and formatted \cite{ma2023query}. An example system architecture for federated prompt conditioning is presented in Appendix I.}

\textbf{Local Knowledge Indexing.} A central challenge in deploying FFMs lies in achieving user-level personalization while safeguarding privacy. Although FMs offer strong generalization, they often falter in highly contextual settings such as personalized document understanding, niche domain tasks, or individual knowledge retrieval, where access to user-specific information is essential~\cite{guan2024federated}. Traditional cloud-based solutions that require uploading user data for fine-tuning or indexing are \textcolor{black}{often constrained by privacy regulation, policy, or bandwidth.} To overcome this, \textit{local knowledge indexing} provides a promising alternative that adheres to the \textit{model to data} principle~\cite{wang2021fieldguide}, where computation occurs locally at the data source.

In RAG frameworks~\cite{lewis2020retrieval}, local indexing enables each client to build and maintain a private, queryable repository of documents, including emails, notes, and clinical records. \textcolor{black}{These documents are embedded into vector representations using a local encoder and stored locally on the client; when stronger protection is required, retrieval and/or generation can be executed inside a Trusted Execution Environment (TEE) (confidential computing), so the plaintext context is only exposed within the protected boundary~\cite{addison2024cfedrag}.} During inference, the retriever performs a similarity search over this private index to retrieve the top-$k$ relevant documents. \textcolor{black}{The retrieved evidence can then be consumed by a generator that is executed locally on the client using globally shared weights, so the private context does not leave the device and the model-to-data principle is preserved. Alternatively, when generation must be offloaded, the generator can be executed within a trusted execution environment at an edge tier, which keeps the retrieved context within a protected boundary~\cite{addison2024cfedrag}.}

Recent systems such as ERAGent~\cite{shi2024eragent} demonstrate \textcolor{black}{that retrieval-augmented assistants can improve task relevance and personalization by leveraging historical interactions (e.g., a memory knowledge base and learned user profiles) alongside external retrieval.} Additionally, modular toolkits such as LlamaIndex have gained popularity for \textcolor{black}{rapidly assembling and experimenting with end-to-end RAG pipelines.} \textcolor{black}{In decentralized settings, C-FedRAG~\cite{addison2024cfedrag} and FRAG~\cite{zhao2024frag} explore secure retrieval across distributed data silos: C-FedRAG leverages confidential computing/TEEs to protect contextual data during orchestration and LLM inference, while FRAG supports encrypted approximate nearest-neighbor search over distributed vector databases using (multi-party) homomorphic encryption.} Altogether, local knowledge indexing can enhance the personalization capabilities of FFMs by anchoring responses in user-owned data, enabling privacy-friendly, adaptive, and domain-specific intelligence. When integrated with federated RAG workflows, it becomes a cornerstone technology for building adaptive PFI.

\textbf{Retriever Adaptation.} 
In federated settings, users often operate across heterogeneous domains such as healthcare, law, or education, \textcolor{black}{and client data are typically non-IID across participants}, which can lead to retrieval mismatches caused by variations in vocabulary, semantics, and document structures. While prompt tuning provides a lightweight mechanism to steer model behavior, it is often insufficient to address domain-specific retrieval errors, particularly when the underlying embedding space of the retriever is misaligned with the user's corpus \textcolor{black}{in a RAG pipeline}~\cite{lewis2020retrieval}. To address this, clients can locally fine-tune only the \textit{retriever} component of the RAG pipeline using parameter-efficient methods such as LoRA~\cite{hu2022lora} and adapter modules~\cite{houlsby2019parameter}. These techniques introduce small, trainable layers into pre-trained retrievers without modifying the backbone model. Because the update size remains minimal, retriever adaptations can be securely aggregated using FL protocols like FedAvg~\cite{mcmahan2017communication} or {secure aggregation}~\cite{bonawitz2017practical}, enabling scalable personalization without compromising privacy or bandwidth efficiency.

In addition, this localized retriever tuning enhances the relevance of retrieved documents while keeping the global generator unchanged, thus \textcolor{black}{preserving the generator’s inference cost and memory footprint (with only a small overhead from the added PEFT modules)}, which are critical requirements for edge deployments~\cite{hu2022lora,houlsby2019parameter}.
\textcolor{black}{Recent in-domain RAG adaptation studies show that retriever-side adaptation can improve retrieval quality and downstream grounded generation, especially under domain shift~\cite{mao2024ragstudio}.}
Moreover, this strategy aligns with privacy-friendly design principles. \textcolor{black}{Because only small adapter/LoRA parameters are communicated (instead of raw data), the update size is bandwidth-efficient. In practice, privacy can be strengthened by protecting these updates with secure aggregation protocols ~\cite{bonawitz2017practical}.}
Appendix J illustrates an example where only the LoRA modules are updated, while all other components remain fixed. The modularity of this setup also allows asynchronous or partial retriever updates across clients~\cite{nguyen2022federatedbuff}, making it suitable for deployment in low-resource or heterogeneous federated environments. In summary, retriever adaptation via LoRA or adapter modules empowers clients to improve retrieval accuracy in a lightweight, private, and communication-efficient manner, enabling scalable deployment of personalized FFMs across diverse domains.

\subsubsection{Server-side Optimization}
While client-side personalization enables tailored responses and privacy-preserving adaptation, global coordination remains essential to ensure fairness, generalization, and robustness across diverse users. In FFMs, the server or regional edge tier is responsible not only for aggregating updates but also for enriching model capabilities through shared knowledge. \textcolor{black}{Importantly, the role of retrieval here is \emph{RAG-inspired} rather than standard inference-time RAG: the server uses retrieval mechanisms to \emph{select}, \emph{route}, or \emph{weight} auxiliary data and training signals that improve coordination across heterogeneous clients, without accessing raw client content.} By leveraging RAG at the server level, we can design intelligent orchestration strategies that optimize training dynamics and enhance sample efficiency without compromising user data. \textcolor{black}{In other words, retrieval is used as a coordination primitive for \emph{optimization} (e.g., task matching, clustering, distillation, curriculum reweighting), rather than solely as an evidence injector for generation.}

\textbf{Task Matching.} In real-world deployments, clients often vary significantly in domain, task complexity, and data availability, ranging from richly annotated clinical corpora to sparse educational notes or niche financial records. To personalize global learning signals without accessing raw client data, {\emph{retrieval-inspired}} task matching \textcolor{black}{is a promising} server-side coordination strategy that enhances contextual alignment during model aggregation. In this paradigm, the server maintains a broad, heterogeneous corpus of publicly available or pseudo-public data, such as open domain texts, instructional datasets, or synthetic knowledge bases. \textcolor{black}{At each communication round, the server utilizes coarse routing signals (e.g., cluster IDs, group-level statistics, or aggregated metadata) to retrieve semantically related examples from this pool that reflect the dominant domain/task (or estimated difficulty) of a \emph{group} of clients}~\cite{ghosh2020efficient}. \textcolor{black}{Unlike standard RAG, which retrieves evidence to condition \emph{inference}}~\cite{lewis2020retrieval}, these retrieved examples can be used to {augment the next round of training} through server-side data-assisted objectives, such as \textcolor{black}{distillation on public/unlabeled data}~\cite{lin2020ensemble,li2019fedmd}, \textcolor{black}{curriculum-style reweighting based on estimated difficulty}~\cite{vahidian2023curricula}, or \textcolor{black}{representation-shaping regularizers (e.g., model-level contrastive alignment)}.

This design introduces a form of implicit personalization by ensuring that each client receives globally aggregated updates that are contextually relevant to its local domain. For example, if {a client cluster} exhibits difficulty in a biomedical classification task, the server retrieves and integrates related medical samples \textcolor{black}{from the public/pseudo-public pool} to reinforce learning during the global update. This targeted augmentation \textcolor{black}{can reduce global--local mismatch, improve optimization stability under non-IID data}, and accelerate convergence \textcolor{black}{when the server-side pool provides adequate coverage of newly emerging domains}. \textcolor{black}{Rather than claiming “fairness by default,” this mechanism is best viewed as improving \emph{relevance} and \emph{sample efficiency}, and it can be paired with explicit fairness objectives when needed}~\cite{li2019fair}. Importantly, since task matching operates over abstract representations (e.g., gradients or embeddings) rather than raw data, it \textcolor{black}{reduces direct exposure but does not by itself guarantee privacy, as gradients can still leak information via inversion attacks \cite{zhu2019deep}}. \textcolor{black}{Consequently, practical deployments often necessitate, stronger privacy requires combining routing logic with secure aggregation and/or client-level differential privacy}~\cite{bonawitz2017practical}. Moreover, the mechanism is especially valuable in open-ended deployments where new clients with previously unseen domains continuously join the federation. In such cases, retrieval-based task matching serves as a lightweight yet effective personalization mechanism, offering adaptive, domain-aware support without requiring direct fine-tuning or explicit supervision.

\textbf{Knowledge Sharing.} In FL, personalization frequently conflicts with global generalization due to statistical heterogeneity across clients. Users may belong to vastly different application domains, such as clinical diagnostics~\cite{raha2024attention}, legal document processing, wireless environments \cite{raha2025security} or educational tutoring, where a single global model often fails to generalize effectively. Cluster-wise knowledge sharing offers a middle ground by grouping clients into \textcolor{black}{clusters with similar training objectives or data distributions}, e.g., \textcolor{black}{update-direction similarity as in FedGroup~\cite{duan2020fedgroup}} or \textcolor{black}{principal angles between client data subspaces as in PACFL~\cite{vahidian2023efficient}}. \textcolor{black}{In text-centric settings, corpus-level similarity (e.g., cosine distance over TF-IDF vectors or latent embeddings) is also a common heuristic when only privacy-safe descriptors are shared.} \textcolor{black}{In a clustered RAG-based design,} each cluster maintains a domain-specific RAG sub-model that comprises a retriever and an adapter module. The retriever is fine-tuned to retrieve documents relevant to the cluster’s specific domain, while the adapter provides lightweight personalization over a shared generator. This design allows intra-cluster model updates to remain focused and semantically aligned. Standard aggregation algorithms such as FedAvg or FedProx~\cite{li2020federated} are used within each cluster to synchronize the sub-models. \textcolor{black}{When exchanging knowledge across clusters, one can avoid sharing raw documents by transferring higher-level signals (e.g., distilled outputs or adapter-level representations), and further strengthen privacy via secure aggregation or differential privacy.} \textcolor{black}{Depending on the system goal, the generator may remain globally shared (optionally updated under the same protections), while retrievers/adapters remain cluster-specific.}

In addition, the RAG framework is particularly valuable in this clustered setting because it decouples knowledge retrieval from language generation. While the generator remains globally shared across all clusters, each cluster-specific retriever ensures that input queries are grounded in domain-relevant knowledge. For instance, the retriever in a biomedical cluster retrieves clinical notes or research abstracts, while the one in a legal cluster fetches case law or statutes. This modularity enables domain-aligned personalization without incurring the computational cost or privacy risk of full-model fine-tuning. Additionally, RAG’s retrieval-based conditioning improves factual grounding, \textcolor{black}{and can mitigate hallucinations when retrieval quality is high}, and enables continual adaptation as the client corpus evolves. Empirical studies support this hybrid strategy. \textcolor{black}{For example, FedGroup~\cite{duan2020fedgroup} clusters clients based on update similarity and achieves notable gains in both accuracy and convergence, improving absolute test accuracy by +14.1\% on FEMNIST over FedAvg~\cite{mcmahan2017communication}, +3.4\% on Sentiment140 over FedProx~\cite{li2020federated}, and +6.9\% on MNIST over FeSEM~\cite{long2023multi}.} Meanwhile, PACFL~\cite{vahidian2023efficient} forms clusters based on principal angles between data subspaces to ensure distributional similarity. \textcolor{black}{In a clustered RAG setting, these ideas can be interpreted as aligning both the client distribution (via clustering) and the contextual evidence used for generation (via cluster-specific retrieval).} In summary, cluster-wise knowledge sharing with RAG enables federated systems to provide domain-specific personalization at scale. By tuning the retriever within each cluster while retaining a globally shared generator, it balances specialization and generalization. This architecture aligns well with the goals of PFI, where personalization, privacy, and communication efficiency must coexist in decentralized and heterogeneous environments.

\textbf{Continual Pre-Training.} Another core limitation of static FMs in FL is their gradual degradation in relevance, particularly in dynamic domains such as healthcare, finance, and education. \textcolor{black}{A typical remedy is to continually pre-train FMs on domain-specific datasets to keep them up-to-date and aligned with evolving data distributions, as illustrated in Appendix K.} However, traditional FL methods often lack mechanisms to integrate evolving knowledge across clients without incurring high communication overhead or violating data privacy. To address this, RAG offers a lightweight and privacy-preserving means of keeping the global generator up-to-date while enhancing personalization across domain-diverse clients.
In RAG-assisted fine-tuning, the server periodically updates the shared generator on either: (a) pseudo-labeled data produced from its own retrieval-augmented pipeline via self-distillation~\cite{izacard2023atlas}, or (b) highly relevant documents dynamically retrieved from curated external sources (e.g., domain-specific databases, medical literature, or trusted web content)~\cite{guu2020retrieval}. These documents are not randomly selected, but semantically aligned with the distribution of prior client tasks or updates, enabling the server to capture emergent knowledge relevant to clusters of users. This improves the generator’s downstream alignment with personalized user contexts, even without direct exposure to raw user data.

The integration of RAG into continual pretraining pipelines can support three key objectives for federated personalization. First, it allows adaptive knowledge updates without relying on centralized annotations or fine-tuning per user, which is crucial in privacy-sensitive environments. Second, it enhances factual grounding by repeatedly training on retrieved hard negatives and diverse contexts, thereby reducing hallucination and domain mismatch. Third, by limiting updates to selectively retrieved, task-relevant content, the approach improves sample efficiency and ensures that model refinement reflects actual user needs rather than generic global trends. This server-side process indirectly fuels personalization at the client level. When a client in a niche domain (e.g., legal research or rare disease diagnosis) queries the generator, the model, fine-tuned on semantically similar retrievals, delivers more accurate, relevant, and context-aware outputs, despite never being fine-tuned on that specific user. Recent studies, such as Few-Shot RAG~\cite{izacard2023atlas}, have shown that this retrieval-centric adaptation not only enhances factual precision but also boosts personalization fidelity in low-resource settings.

\textcolor{black}{
Although retrieval-augmented continual pretraining can improve freshness and grounding, it can also introduce \emph{retrieval-induced drift} when the retrieval store evolves over time. This drift arises when imperfect, biased, stale, or adversarially influenced evidence is repeatedly injected into the context during inference and/or reused as training signal during continual updates, gradually amplifying spurious correlations and increasing hallucination and overconfident generations if unreliable or conflicting evidence is not detected and handled. Moreover, poisoning the retrieval corpus can induce \emph{retrieval backdoors} that selectively steer downstream outputs while leaving the base model unchanged, highlighting the need for security-aware corpus maintenance~\cite{xue2024badrag}. 
To directly address these risks, robust \emph{content curation} is essential for long-running RAG systems: practical mechanisms include provenance-aware indexing with source vetting/whitelisting, versioning and temporal validity checks for evolving documents, de-duplication and conflict detection to avoid repeatedly reinforcing erroneous claims, and continuous auditing/traceback to identify and quarantine suspicious or poisoned passages~\cite{zhang2025traceback}.}

\textcolor{black}{
A widely adopted mitigation direction is to make \emph{retrieval-quality awareness} explicit \emph{before} generation and \emph{before} any model update. Corrective RAG (CRAG) methods estimate retrieval quality and trigger corrective actions, such as query rewriting, additional retrieval, or selective context recomposition, when evidence is unreliable, as exemplified by CRAG and RQ-RAG~\cite{yan2024crag,chan2024rqrang}. Complementarily, self-reflective RAG approaches encourage models to decide when retrieval is needed and to critique evidence and generations, improving faithfulness under noisy retrieval (e.g., SELF-RAG, SimRAG, and Self-BioRAG)~\cite{asai2024selfrag,xu2025simrag,jeong2024selfbiorag}. In our framework, these safeguards naturally extend to continual pretraining by \emph{gating} updates on high-confidence evidence and by excluding low-trust or conflicting contexts identified by the curation and quality-checking stages. We summarize the key RAG-enabled technologies for personalized federated intelligence in Appendix L.}

\textcolor{black}{Table~\ref{tab:technique_comparison} summarizes how major personalization mechanisms trade off communication overhead, personalization fidelity, and compatibility with heterogeneous clients. In practice, \emph{prompt tuning} is most effective in large-scale cross-device deployments where uplink bandwidth and on-device compute are tightly constrained, and personalization mainly targets format/style/intent control. \emph{LoRA} is preferable when clients exhibit stronger non-IID domain/task shifts and can afford lightweight local backpropagation, as it offers higher behavioral fidelity by updating internal representations while keeping payloads small. \emph{RAG} is most effective when personalization is driven by user-owned or rapidly changing knowledge; local indexing and retrieval improve grounding and freshness, and can be combined with PEFT (e.g., LoRA or prompt tuning) when both knowledge and behavioral/style adaptation are required.}

\textcolor{black}{Note that, although significant efforts have been made to explore the core stages of the PFI pipeline, including efficient personalization, trustworthy adaptation, and adaptive refinement via RAG, we acknowledge that \cite{zhang2025personalization} also offers a complementary perspective on personalizing FMs through various approaches, such as prompting methods, representation learning, and RLHF. These techniques, developed from different perspectives, can also improve the design of personalized FL frameworks capable of accommodating individual user preferences.}

\begin{table}[t]
\centering
\scriptsize
\textcolor{black}{\caption{\textcolor{black}{Comparative analysis of key personalization mechanism in PFI.}}
\label{tab:technique_comparison}
\vspace{-10px}
\setlength{\tabcolsep}{2pt}
\renewcommand{\arraystretch}{1.15}
\setlength{\emergencystretch}{1em}
\begin{tabularx}{\linewidth}{P{0.85cm} P{2.2cm} P{2cm} P{2.2cm} P{2.3cm} Y}
\toprule
\textbf{\textcolor{black}{Tech.}} & \textbf{\textcolor{black}{What is updated}} & \textbf{\textcolor{black}{Comm. overhead}} & \textbf{\textcolor{black}{Fidelity}} & \textbf{\textcolor{black}{Hetero. compat.}} & \textbf{\textcolor{black}{Most effective scenario}} \\
\midrule
\textcolor{black}{LoRA} &
\textcolor{black}{Trainable low-rank adapter matrices inserted into selected projections (e.g., attention/MLP); base weights frozen} &
\textbf{\textcolor{black}{Low}} \textcolor{black}{(communicate only adapter deltas; size scales with rank and chosen layers)} &
\textbf{\textcolor{black}{High}} \textcolor{black}{(modifies internal representations; strong for domain/task shift vs pure prompting)} &
\textcolor{black}{\textbf{Medium} (naive averaging assumes identical adapter shapes; hetero ranks/layers need padding, clustering, or separate heads)} &
\textcolor{black}{Non-IID clients where you need \emph{behavioral/domain} adaptation with limited uplink; devices can afford light backprop on a few layers} \\
\midrule
\textcolor{black}{Prompt} &
\textcolor{black}{Trainable soft prompts / prefix vectors (optionally layer-wise prefix tuning); base model frozen} &
\textcolor{black}{\textbf{Very low} (only a small set of prompt vectors; minimal round payload)} &
\textcolor{black}{\textbf{Medium} (excellent for format/style/task cues; limited capacity for large distribution shifts)} &
\textcolor{black}{\textbf{High} (tiny memory/compute footprint; robust to device diversity and partial participation)} &
\textcolor{black}{Massive cross-device settings with tight uplink/compute where quick personalization is needed (e.g., style, tone, and intent) without deep model changes} \\
\midrule
\textcolor{black}{RAG} &
\textcolor{black}{User/private corpus $\rightarrow$ embeddings/index; retrieval pipeline; generator often shared/frozen (retriever may be updated periodically)} &
\textcolor{black}{\textbf{Low--Medium} (low if indexing stays local; higher if sharing/training retriever or syncing embeddings)} &
\textcolor{black}{\textbf{High for knowledge} (grounding, freshness, and source control), \textbf{medium for style} unless paired with PEFT} &
\textcolor{black}{\textbf{Medium--High} (depends on storage/retrieval support; can be local, edge-assisted, or clustered)} &
\textcolor{black}{When personalization is mainly \emph{user-owned or rapidly changing knowledge} and reducing hallucinations/keeping answers up-to-date matters; add PEFT if tone/behavior must also adapt} \\
\bottomrule
\end{tabularx}
\vspace{-20px}}
\end{table}

\subsection{Summary and Lessons Learned}

This section has examined how RAG frameworks can support adaptive PFI by enabling lightweight, privacy-preserving, and domain-aligned personalization across client-side and server-side designs, without requiring full-model retraining or raw data exchange.

\textbf{Takeaways:}
RAG enables effective personalization through both client-side adaptation (e.g., prompt conditioning and local retrieval) and server-side coordination (e.g., task matching and cluster-wise knowledge sharing). Its modular design, which decouples retrieval from generation, allows efficient and flexible adaptation in federated settings.

\textbf{Connection to PFI:}
RAG naturally aligns with the core principles of PFI, including personalization, privacy preservation, and communication efficiency. By supporting per-user or per-cluster adaptation under the \textit{model-to-data} paradigm, RAG provides a practical foundation for scalable and adaptive federated intelligence. Open challenges include balancing personalization and generalization across heterogeneous users, ensuring efficiency and fairness at scale, and designing privacy-preserving retrieval mechanisms suitable for resource-constrained devices. In addition, continual server-side adaptation must carefully manage retrieved content to avoid model drift and hallucination.

\section{Future Directions}
\label{sec_future}
In the following section, we discuss promising future directions, including Meta-PFI, quantum-enabled PFI, and sustainable PFI.

\subsection{Meta-PFI}
Integrating autonomous capabilities into PFI opens new avenues for enhancing adaptability and intelligence. We term this emerging paradigm as meta-personalized federated intelligence (Meta-PFI). Meta-PFI operates within a dynamic user-agent-world interaction loop: users (human or avatar) provide real-time behavioral data and feedback; agents (embodied federated agents) autonomously adapt strategies based on user input or environmental perception; and the world (physical or virtual) shapes both data collection and learning dynamics. 

\textcolor{black}{To enable context-aware federated adaptation, virtual and mixed-reality environments (VR, AR, MR, XR), as well as digital twins, can serve as structured and immersive ecosystems for interaction and data generation~\cite{aloqaily2022integrating}. These environments facilitate the collection of fine-grained user feedback and the development of rich agent perception, allowing personalized insights from diverse users and contexts to be integrated into the global FM without sharing raw data. Specifically, in this paradigm, embodied federated agents form perception–action loops within virtual or physical environments: they locally observe user interactions (e.g., gaze, gestures, and speech) and environmental dynamics through these immersive technologies, fine-tune their local FMs for personalized reasoning, and periodically transmit model updates instead of raw data to a central aggregator. This mechanism enables coordinated and privacy-preserving learning across heterogeneous users while maintaining personalization and adaptability. Recent advances in embodied AI research provide concrete empirical grounding for this direction. For example, Habitat 3.0~\cite{puighabitat} provides a large-scale embodied simulation environment that enables realistic perception–action loops for training and evaluating embodied agents. RoboCat~\cite{bousmalisrobocat} further demonstrates how foundation models can be adapted into robotic and embodied systems through multi-task and multi-embodiment learning, achieving rapid adaptation to novel tasks using only a small number of target examples. Overall, these systems exemplify how embodied agents can continuously learn from multimodal sensory data, which, to some extent, aligns with Meta-PFI’s vision of using federated adaptation pipelines to connect user-specific interactions with FM refinement.}

\subsection{Quantum-Enabled PFI}

Quantum-enabled PFI offers a promising alternative by leveraging quantum parallelism, cryptography, and machine learning to enhance optimization, security, and efficiency. Quantum-secure hardware, such as quantum mask circuits~\cite{ain2025secure}, can serve as a powerful hardware component in the context of PFI. By integrating these circuits into edge devices within PFI, sensitive elements-like user data, model parameters, and personalization layers-are protected against side-channel attacks, including power analysis, electromagnetic emission, and timing attacks, which are especially threatening in distributed, heterogeneous environments. Overall, unlike traditional algorithmic defenses (e.g., differential privacy or secure aggregation), quantum-enabled circuits~\cite{ju2024quantum} offer hardware-level protection that complements software-based methods. This hybrid approach ensures end-to-end robustness, enhancing both the privacy and trustworthiness of PFI during training and inference.

To address vulnerabilities associated with centralized aggregation in PFI, it is crucial to rethink server and client selection strategies in a quantum-aware context \cite{mathur2025federated}. A promising direction involves the development of a quantum-probabilistic framework~\cite{chowdhury2019probabilistic} that leverages quantum randomness and quantum interference to enable secure, unpredictable, and highly dynamic aggregation processes. In this approach, quantum circuits can generate secure random selection vectors to enable decentralized, entropy-driven aggregation. This may mitigate server collusion and biased sampling while enhancing resilience and eliminating single points of failure in traditional centralized designs. \textcolor{black}{However, the limitation of quantum hardware in terms of qubit count or low gate fidelity prevents the direction from large-scale model aggregation. To overcome this challenge, hybrid quantum-classical machine learning approaches have been proposed in \cite{chen2021federated}, where a classical pre-trained model is used to map the input image to a much lower dimension before it is fed into the quantum circuit for classification. Similarly, authors in \cite{cowlessur2025hybrid} also proposed the hybrid model to accommodate a larger number of clients within the network. On the other hand, the authors in \cite{10651123} designed a 4-qubit quantum neural network and trained it using the FL framework, which still achieves a relatively high performance. Furthermore, the work conducted privacy-preserving aggregation algorithms and quantum gates for encryption to further protect client data. Another approach to resolve the hardware constraint is variational quantum algorithms, which have emerged as the leading strategy to obtain quantum advantage on Noisy Intermediate-Scale Quantum (NISQ) computers \cite{cerezo2021variational}.} Another promising direction is the integration of quantum wireless networks~\cite{wang2022quantum}, enabling ultra-secure, low-latency communication across distributed clients and servers. Leveraging technologies such as entanglement distribution and quantum repeaters can extend the distance of quantum communication. Additionally, integrating quantum wireless communication with edge intelligence and context-aware networking would enhance PFI's robustness and efficiency, facilitating secure AI development and beyond ecosystems. \textcolor{black}{Regardless of these promising advancements, it faces a couple of limitations before deploying in real-world applications, such as the fragility of communication qubits, the stochastic success probability of entanglement swapping in quantum networks, and also the lack of a fully functional physical quantum repeater~\cite{kumar2025quantum}. Additionally, the security enhancement from cryptographic and quantum techniques can lead to an increase in computational resources and make it infeasible for resource-constrained edge devices, which motivates the shift in focus toward the hybrid post-quantum cryptography research \cite{turnip2025towards}.}

\subsection{Sustainable PFI}
Sustainable PFI represents a promising paradigm at the intersection of personalized foundation models and environmental sustainability, emphasizing energy-aware model optimization, carbon-aware scheduling, and the utilization of renewable energy.

\textcolor{black}{To ensure scalable and environmentally responsible deployment of PFI at the edge, energy-aware model optimization is essential, especially as PFI scales across heterogeneous devices with diverse computational and battery capacities~\cite{li2022energy}. For instance, \cite{husom2025sustainable} measures the energy consumption of quantized LLMs running on a low-resource edge device and demonstrates that even the inference of LLMs on edge hardware can incur substantial energy and latency costs, highlighting the real energy burden of on-device LLM deployment. Similarly, \cite{xu2025camel} shows that balancing GPU frequency and batch size can reduce the energy-delay product by 12.4\%–29.9\% on an embedded platform, underlining the need for energy-efficient configurations when running LLMs at the edge. Moreover, combining model optimization with carbon-aware scheduling can further reduce the carbon footprint. For instance, GreenScale~\cite{kim2023greenscale} introduces a carbon-aware scheduling framework that accounts for the time and location-based carbon intensity of edge-cloud infrastructures, enabling optimized task placement for applications such as AR and VR, and reducing carbon emissions by up to 29.1\%. In addition, renewable energy integration in sustainable PFI can involve leveraging solar or wind-powered edge devices to enhance green energy availability, allowing for the planning of training rounds in advance. For example, \cite{nguyen2023optimal} examines renewable energy-powered edge clouds and demonstrates how workload allocation across distributed edge sites can reduce operating costs and carbon emissions while maintaining quality of service. These studies collectively highlight the importance of designing personalized models that are environmentally sustainable, motivating future research on adaptive lightweight architectures, carbon-aware scheduling, and renewable energy utilization.}

\section{Conclusion}
\label{sec:conclusion}

\textcolor{black}{This survey envisions artificial personalized intelligence (API) as a natural complement to artificial general intelligence (AGI), emphasizing intelligence that is inherently user-centric, context-aware, and continuously adaptive. Within this landscape, personalized federated intelligence (PFI) emerges as a foundational paradigm, providing a principled framework for delivering large model capabilities to edge environments while preserving personalization and privacy.} \textcolor{black}{By revisiting the foundations of FL and FMs, we have proposed unified taxonomies that illuminate their structural components and the opportunities created by their convergence. Through the three stages of the PFI pipeline, we have synthesized: (1) efficiency-driven methods that reduce computation and communication cost through model compression and parameter-efficient fine-tuning, (2) trust-oriented mechanisms covering bias mitigation, fairness, interpretability, hallucination reduction, and robustness, and (3) adaptive personalization techniques that leverage user heterogeneity and contextual signals through lightweight, retrieval-driven, and modular adaptation mechanisms. Together, these insights clarify how PFI can serve as a compelling pathway towards the emerging era of API. Looking ahead, we outline several promising directions, including Meta-PFI, quantum-enabled learning, and sustainable intelligence. Beyond the scope of PFI, we believe that other research communities will likewise play a pivotal role in shaping the evolution of API, which opens up intriguing opportunities for future research.}

\bibliographystyle{ACM-Reference-Format}
\bibliography{bib_local_clean}

\appendix

\end{document}